\documentclass[onefignum,onetabnum]{siamonline171218}

\usepackage{array}
\usepackage{stfloats}
\usepackage{url}
\usepackage{verbatim}
\usepackage{subfigure}%
\usepackage{multirow}%
\usepackage{amsmath}
\usepackage{amssymb}
\usepackage{mathrsfs}%
\usepackage{mathrsfs}%
\usepackage{xcolor}%
\usepackage{textcomp}%
\usepackage{manyfoot}%
\usepackage{booktabs}%
\usepackage{listings}%
\usepackage{adjustbox}
\usepackage{diagbox}
\usepackage{bm}%
\usepackage{bbding}%
\usepackage{cite}
\usepackage{float}

\usepackage{lipsum}
\usepackage{amsfonts}
\usepackage{graphicx}
\usepackage{epstopdf}

\usepackage{algorithmic}
\ifpdf
  \DeclareGraphicsExtensions{.eps,.pdf,.png,.jpg}
\else
  \DeclareGraphicsExtensions{.eps}
\fi

\newsiamremark{remark}{Remark}
\newsiamremark{hypothesis}{Hypothesis}
\crefname{hypothesis}{Hypothesis}{Hypotheses}
\newsiamthm{claim}{Claim}

\headers{Anisotropic Diffusion Probabilistic Model}{Jingyu Kong, Yuan Guo, Yu Wang, and Yuping Duan}

\title{Anisotropic Diffusion Probabilistic Model for
Imbalanced Image Classification\thanks{Submitted to the editors DATE.
}}

\author{Jingyu Kong\thanks{Center for Applied Mathematics,Tianjin University.}
\and Yuan Guo\footnotemark[2]
\and Yu Wang\thanks{DAMO Academy, Alibaba Group.}
\and Yuping Duan\thanks{School of Mathematical Sciences, Beijing Normal University (\email{ypduan@bnu.edu.cn}).}}

\usepackage{amsopn}

\makeatletter
\newcommand*{\addFileDependency}[1]{
  \typeout{(#1)}
  \@addtofilelist{#1}
  \IfFileExists{#1}{}{\typeout{No file #1.}}
}
\makeatother

\newcommand*{\myexternaldocument}[1]{%
    \externaldocument{#1}%
    \addFileDependency{#1.tex}%
    \addFileDependency{#1.aux}%
}

\myexternaldocument{ex_supplement}

\begin{document}

\maketitle

\begin{abstract}
Real-world data often has a long-tailed distribution, where the scarcity of tail samples significantly limits the model's generalization ability. Denoising Diffusion Probabilistic Models (DDPM) are generative models based on stochastic differential equation theory and have demonstrated impressive performance in image classification tasks. However, existing diffusion probabilistic models do not perform satisfactorily in classifying tail classes. In this work, we propose the Anisotropic Diffusion Probabilistic Model (ADPM) for imbalanced image classification problems. We utilize the data distribution to control the diffusion speed of different class samples during the forward process, effectively improving the classification accuracy of the denoiser in the reverse process. Specifically, we provide a theoretical strategy for selecting noise levels for different categories in the diffusion process based on error analysis theory to address the imbalanced classification problem. Furthermore, we integrate global and local image prior in the forward process to enhance the model’s discriminative ability in the spatial dimension, while incorporate semantic-level contextual information in the reverse process to boost the model’s discriminative power and robustness. Through comparisons with state-of-the-art methods on four medical benchmark datasets, we validate the effectiveness of the proposed method in handling long-tail data. Our results confirm that the anisotropic diffusion model significantly improves the classification accuracy of rare classes while maintaining the accuracy of head classes. On the skin lesion datasets, PAD-UFES and HAM10000, the F1-scores of our method improved by 4\% and 3\%, respectively compared to the original diffusion probabilistic model.
\end{abstract}



\section{Introduction}\label{introduction}
The class imbalance refers to the phenomenon where the number of samples is unevenly distributed across different categories, with some categories (majority classes) having significantly more samples than others (minority classes), which is particularly common in fields such as medical imaging \cite{krawczyk2016learning,shen2015long}, natural scenes \cite{sinha2022class,ma2024geometric} and security surveillance \cite{liu2019large}, etc. 
In medical image classification, the incidence rates of different diseases vary significantly, and accurately predicting rare diseases holds great clinical importance. However, class imbalance can cause classification models to favor majority classes, resulting in poorer performance when recognizing minority classes and reducing overall classification accuracy.
 
Numerous strategies have been extensively explored to address the long-tail problems. These can be broadly categorized into three types: data-based methods, model-based methods, and effective loss functions.
Data-based methods typically involve techniques like oversampling the minority classes, undersampling the majority classes, or synthesizing new examples for the minority classes to balance the dataset. The resampling methods entail either oversampling minority classes \cite{chawla2002smote,han2005borderline} or undersampling majority classes \cite{jeatrakul2010classification}. Data synthesis methods generate new samples to augment the representation of minority categories within the dataset, thereby mitigating imbalanced data issues \cite{10.1007/978-3-030-65414-6_9,galdran2021balanced}. Transfer learning is further categorized into feature transfer and knowledge transfer. The former involves applying feature representations learned from a source domain to a target domain and extracting relevant data features within the target domain \cite{park2022majority,wang2017learning}. The latter involves leveraging knowledge, rules, or prior information acquired from the source domain to aid the learning process of the target task \cite{jamal2020rethinking,liu2019large}. Model-based methods, on the other hand, focus on modifying the model architecture or training process to better handle the class imbalance. It can involve using different models for different classes or adjusting the model's complexity based on the class distribution. Notably, metric learning endeavors to precisely model the boundaries of minority-class data by acquiring a suitable metric function and assessing the similarity or dissimilarity between samples within the data space \cite{yang2020rethinking,zhang2017range}. Feature reconstruction methods aim to map the original feature space to a more discriminative feature space through the utilization of a feature reconstruction network, thereby enhancing the accuracy of minority-class data classification within the transformed feature space \cite{zhou2020bbn}. 
He \emph{et al.} \cite{doi:10.1137/22M1503579} developed a signal transformation method inspired by the fruit fly olfactory system through biomimicry, achieving data sparsification and enhancing the robustness in handling data with random errors. 
Last but not least, adjustments to the loss function are aimed at either imposing higher penalties for misclassifying minority classes or modifying the contribution of each class to the overall loss based on its frequency in the dataset. By utilizing a weighted loss function, it allows the model to prioritize samples from minority categories during training by assigning varying weights to samples from different categories \cite{lin2017focal,9848833}. The well-known focal loss \cite{lin2017focal} was proposed to reduce the weight of easy samples and emphasize attention toward challenging samples. The Tversky loss \cite{salehi2017tversky} was introduced for the imbalanced image segmentation problems, which used adjustable weights for false positives and false negatives. Jin, Lang and Lei \cite{jin2023optimal} proposed adaptive bias loss based on the optimal transport theory to shift the label distribution in the class-imbalanced trainin, which does not depend on the category frequencies in the training set and also can avoid the overfitting of tail classes. Yang \emph{et al.} \cite{9502525} introduced a novel approach for handling imbalanced multiclass classification by optimizing the multiclass AUC through an empirical surrogate risk minimization framework. 

\begin{figure*}[t]
    \centering
    \includegraphics[scale=0.35]{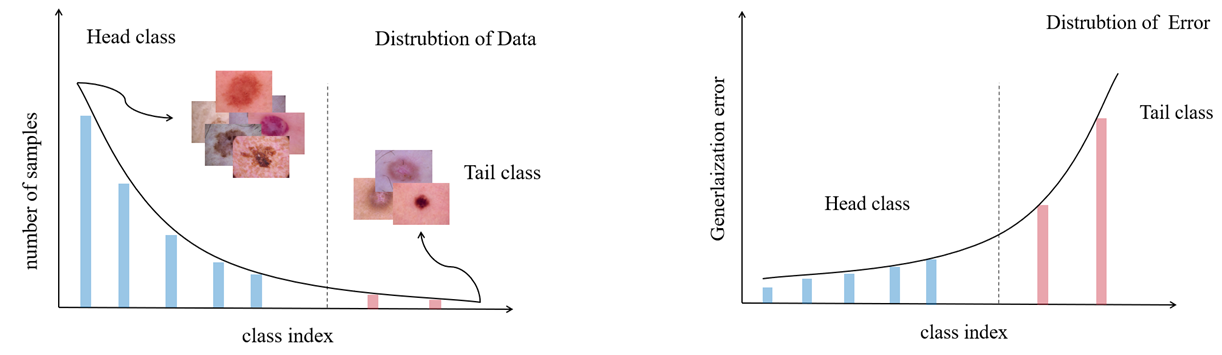}
    \caption{The generalization error of the classification model, where the tail classes contribute more to the generalization error.}
    \label{motivation1}
\end{figure*}

The randomness plays an important role in addressing class imbalance tasks, where class distribution can be adjusted through random sampling \cite{chawla2002smote,han2005borderline} and random data augmentations \cite{he2009learning}. 
Denoising Diffusion Probabilistic Models (DDPMs) \cite{ho2020denoising} are a class of generative models that use stochastic differential equations (SDEs) to model the data distribution. The inherent design of DDPMs introduces controlled randomness through the gradual addition of Gaussian noise during the diffusion process. The stochasticity is pivotal as it enables the model to explore various states within the data distribution, enhancing its capacity to generalize across different sample types. The model learns to reverse such process, iteratively denoising the data, converging to the target distribution through a series of learned steps \cite{sohl2015deep, song2019generative}. The randomness inherent in DDPMs is particularly beneficial in addressing the challenge of class imbalance. By introducing noise in a controlled manner, DDPMs can generate synthetic samples that mimic under-represented classes, thus artificially balancing the dataset. The powerful generative capabilities are critical important in scenarios such as medical image processing tasks, where rare diseases may be underrepresented in the available datasets. Through strategic noise manipulation, DDPMs ensure that every class is adequately represented in the training process, improving the robustness and accuracy of the model across different classes \cite{nichol2021improved, jain2016structural}. Therefore, DDPM has been successfully applied to various image processing tasks, including medical image segmentation \cite{wu2024medsegdiff, wu2024medsegdiffv2}, image super-resolution \cite{ho2022cascaded}, text to image generation \cite{ramesh2022hierarchical,saharia2022photorealistic}, etc.
Beyond these applications, DDPMs and their variants have also been tailored to specifically tackle long-tail and class imbalance problems across various domains. For instance, researchers have developed variants of DDPM that adjust the diffusion process according to the rarity of the class, thereby giving more representation to rarer classes. These modified DDPMs have been successfully applied in fields ranging from natural language processing to complex image generation, demonstrating significant improvements in dealing with imbalanced datasets. Frisch \emph{et al.} \cite{frisch2023synthesising} employed a conditional generative model based on Denoising Diffusion Implicit Models and Classifier-Free Guidance to synthesize high-quality examples tailored to the underrepresented phases and tool combinations in cataract surgeries. For graph outlier detection, Liu \emph{et al.} \cite{liu2023data} introduced the Graph Outlier Detection Model, which leverages latent diffusion models to enhance the performance of supervised graph outlier detection tasks, particularly in the context of class imbalance. Han \emph{et al.} \cite{han2024latent} introduced a method addresses the long-tailed distribution problem by generating pseudo-features in the latent space using a Denoising Diffusion Implicit Model. The process involves encoding an imbalanced dataset into features, training a diffusion model on these features, and using the generated pseudo-features alongside real encoded features to train a classifier. Qin \emph{et al.} \cite{qin2023class} proposed Class-Balancing Diffusion Models which aims to improve the generation quality on tail classes by adjusting the conditional transfer probability during sampling. By incorporating a distribution adjustment regularizer, the model implicitly enforces a balanced prior distribution during the sampling process.

\begin{figure*}[t]
    \centering
    \includegraphics[scale=0.365]{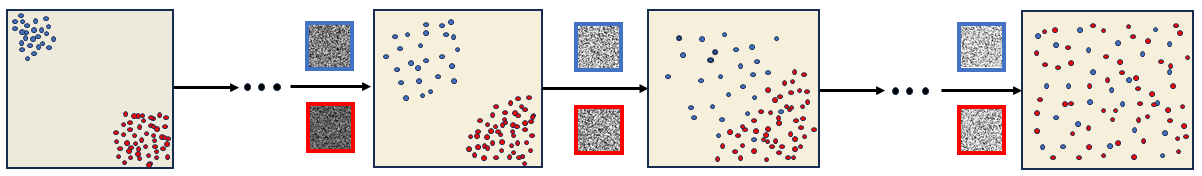}
    \caption{The motivation behind our framework is to apply anisotropic noise to different classes during the diffusion process, effectively balancing the diffusion speed between head and tail classes.}
    \label{motivation2}
\end{figure*}

In this paper, we propose an Anisotropic Diffusion Probabilistic Model (ADPM) that enhances the diffusion speed of tail class samples in the feature space by increasing their randomness. The lower noise levels in head classes direct the model to focus on stable features, while the higher noise in tail classes encourages the model to find stable classification boundaries amid greater variability. As shown in Figure \ref{motivation2}, by adjusting the noise distribution to control diffusion speed, introducing larger noise to tail classes can generate samples and improve classification accuracy. More specifically, we introduce anisotropic noise based on the inherent characteristics of the data distribution, effectively reducing the model's bias toward head classes while enhancing its generalization ability for tail classes. As a result, our method shows promise in handling complex distributions and capturing inherent patterns in medical data. Additionally, we integrate effective prior information into the classification model by embedding image features, thereby improving its ability to discriminate across spatial dimensions and incorporating semantic-level contextual information, which enhances the model's discriminative power and robustness. 
Our experimental results demonstrate the superior performance of our approach on multiple benchmark datasets, including PAD-UFES, HAM10000, SCIN and Hyper-Kvasir. Furthermore, our method improves the classification accuracy of rare classes, proving the effectiveness of combining anisotropic noise and advanced feature fusion techniques in handling imbalanced medical image classification tasks. 

To sum up, our contributions can be summarized as follows:
\begin{itemize}
    \item We propose an anisotropic diffusion probabilistic model for image classification tasks, which leverages randomness to alleviate data distribution imbalance, thereby improving the prediction performance of the diffusion probabilistic model in class imbalance classification problems. 
    \item Based on the generalization error theory, we propose an imbalance-sensitive generalization error bound to guide the noise distribution of different classes in diffusion probabilistic models. Our imbalance-sensitive noise distribution can balance the distribution of head and tail classes, thereby improving the classification performance in class imbalance problems.
    \item We evaluate the effectiveness of our ADPM on different medical image classification tasks, including a dataset of clinical skin images, a dataset of dermatoscopic images, and a dataset of the gastrointestinal tract. The results demonstrate that our approach can significantly improve the classification accuracy of the minority categories.
\end{itemize}

\section{Anisotropic diffusion probabilistic model for classification}\label{anisodiff}
\subsection{Notations and preliminaries}
Let $S =\{\mathbf{x}_i, \mathrm{y}_i\}_{i=1}^n$ denote a labeled dataset with $n\in \mathbb Z_{>0}$ samples. We define an input $\mathbf{x}_i\in\mathbb R^d$ and its corresponding label $\mathrm{y}_i\in\{1,2,\ldots,k\}$.  Let $\mathcal{S}_j=\{\mathbf x_i\}_{i=1}^{n_j}$ be the samples belonging to class $j$ and $n_j\in \mathbb Z_{>0}$ be the number of examples in class $j$. 
We use $p(\mathbf{x}|\mathrm{y})$ to denote the class-conditional distribution, and assume the class-conditional distribution is the same at the training and testing set. For class balanced dataset, the class frequencies are equal for the fairness among categories as follows:
\begin{equation*}
\displaystyle\int p(\mathbf{x}| \mathrm{y} = i)d\mathbf{x} = \int p(\mathbf{x}|\mathrm{y}=j)d\mathbf{x}, \quad \forall i,j.
\end{equation*}
On the other hand, we can assume there is the following relationship for the long-tailed dataset
\begin{equation*}
\left\{\begin{array}{ll}
\displaystyle\int p(\mathbf{x}| \mathrm{y} = i)d\mathbf{x}\geq \int p(\mathbf{x}|\mathrm{y}=j)d\mathbf{x}, ~~ \forall i\leq j,\\
\displaystyle \lim\limits_{i\rightarrow k} \int p(\mathbf{x}| \mathrm{y}=i) d\mathbf{x} =0.
\end{array}\right.
\end{equation*}
Note that the class volumes decay succesively with the ascending class indexes and finally approach zero in the last few classes. 

\subsection{Denoising Diffusion Probabilistic Model}
Han \emph{et al.} \cite{han2022card} introduced classification and regression diffusion models, which combine a denoising diffusion-based conditional generative model and a pre-trained conditional mean estimator, to accurately predict the distribution of response variable $\mathbf{y}$ given its covariates $\mathbf{x}$. Both forward and reverse processes in diffusion model are parameterized Markov chains, with the reverse process used to generate real data. The forward prior conditional distribution is
\begin{equation}\label{e1}
    q\left(\mathbf{y}^{t} \mid \mathbf{y}^{0}\right)=N\left(\mathbf{y}^{t} ; \sqrt{\overline{\alpha}^{t}} \mathbf{y}^{0},\left(1-\overline{\alpha}^{t}\right) \mathbf{I}\right),
\end{equation}
where $\overline{\alpha}^{t}=\prod_{i=1}^t\beta^i$ and $\beta^i$ denotes the intensity of the noise added to the image at time $i$.

Specifically, for the classification tasks, the forward process entails progressively introducing Gaussian noise to class label $\mathbf{y}^0$ over a series of $T$ noisy steps, gradually accumulating Gaussian noise until the data becomes completely random via
\begin{equation}\label{e2}
    \mathbf{y}^{t}=\sqrt{\overline{\alpha}^{t}} \mathbf{y}^{0}+\sqrt{\left(1-\overline{\alpha}^{t}\right)} \bm{\epsilon},
\end{equation}
where $\bm{\epsilon}$ denotes the standard Gaussian distribution used during reparameterized sampling, i.e., $\bm{\epsilon} \sim N(\mathbf{0}, \mathbf{I})$. 
In the reverse process, a neural network is trained to generate clean data by removing noise, starting from $p(\mathbf{y}^T)$, which can be formally expressed as follows:
\begin{equation}\label{e3}
p_{\theta}\left(\mathbf{y}^{0: T}\right)=p\left(\mathbf{y}^{T}\right) \prod_{t=1}^{T} p_{\theta}\left(\mathbf{y}^{t-1} \mid \mathbf{y}^{t}\right),
\end{equation}
with 
\begin{equation*}\label{e4}
p_{\theta}\left(\mathbf{y}^{t-1} \mid \mathbf{y}^{t}\right)=N\left(\mathbf{y}^{t-1}; \mathbf{\mu}_{\theta}\left(\mathbf{y}^{t}, t\right), \mathbf{\Sigma_{\theta}}\left(\mathbf{y}^{t}, t\right)\right),
\end{equation*}
and $p(\mathbf{y}^T)=N(\mathbf{y}^T;\mathbf{0},\mathbf{I})$ and $\theta$ denoting the network parameters. The sampling process is straightforward, Gaussian noise is sampled from $p(\mathbf{y}^T)$, and the reverse denoising process distribution $p_{\theta}\left(\mathbf{y}^{t-1} \mid \mathbf{y}^{t}\right)$ is iteratively run for $T$ steps to generate a new class label. More specifically,  the reverse iteration is given as follows:
\begin{equation}\label{e5}
\mathbf{y}^{t-1}=\frac{1}{\sqrt{\alpha^t}}\left(\mathbf{y}^t-\frac{1-\alpha^t}{\sqrt{1-\overline{\alpha}^t}}{\bm{\epsilon}_\theta}(\mathbf{y}^t,t)\right)+\sigma^t\mathbf{z},
\end{equation}
with $\mathbf{z}\sim N(\mathbf{0},\mathbf{I})$. We are thus able to estimate the clean image by the pretrained denoiser ${\bm{\epsilon}_\theta}(\mathbf{y}^t,t,\mathbf{x})$.

\subsection{Anisotropic Diffusion Probabilistic Model} The original diffusion model utilizes isotropic speeds to diffuse different images, achieving good results in tasks such as image generation, reconstruction, and segmentation. For the problem of class-imbalanced image classification, due to the biased distribution of images, we propose using different diffusion speeds for different categories of images, i.e., the anisotropic diffusion probabilistic model. Specifically, by increasing the diffusion speed of tail classes, we aim to mitigate classification errors caused by class imbalance during the diffusion process. This tailored approach allows for dynamic adjustment of diffusion parameters based on the specific needs of different image categories, thus enhancing the model's sensitivity to minority classes. By differentiating the diffusion speeds, the model not only addresses the inherent biases in the training data but also enhances the overall accuracy and robustness of the classification system. Let $\mathcal{S}$ be a dataset composed of $k$ classes, i.e., $\mathcal{S}=\bigcup_{j=1}^k \mathcal{S}_j$. Among the classes, different noise levels are employed to control the diffusion speed in forward process, i.e., $\bm\lambda=(\lambda_1,\ldots,\lambda_k)$, where $\lambda_j$ is the noise level variable for the $j$th class. Then the prior distribution becomes as follows:
\begin{equation*}\label{e8}
q\left(\mathbf{y}^{t}_{j} \mid \mathbf{y}^{t-1}_{j}\right):=N\left(\mathbf{y}^{t}_{j}; \sqrt{1-\lambda_j\beta^{t}} \mathbf{y}^{t-1}_{j}, \lambda_j\beta^{t} \mathbf{I}\right),
\end{equation*}
and 
\begin{equation*}\label{e9}
q(\mathbf{y}_j^{1:T}|\mathbf{y}_j^0)=\prod_{t=1}^Tq\Big(\mathbf{y}_j^t|\mathbf{y}_j^{t-1}\Big),
\end{equation*}
for the $j$th class. For sampling, from the conditional distribution, we have 
\begin{equation}\label{e10}
\mathbf{y}_j^{t}=\sqrt{\prod_{i=1}^{t}\left(1-\lambda _j\beta^{i}\right)} \mathbf{y}_j^{0}+\sqrt{1-\prod_{i=1}^{t}\left(1-\lambda _j\beta^{i}\right)} \bm \epsilon.
\end{equation}
In fact, it can be observed from \eqref{e10} that for the tail-end classes, stronger levels of noise are added, which can effectively provide stronger regularization for classes with fewer samples.  Therefore, the denoising network has to employ different diffusion speeds to estimate the noise functions of different classes at the same time $t$, which can ensure the classification accuracy of tail-end classes and reduce the impact of class imbalance issues. Similar to original DDPM, once the denosing network training is complete, the sampling process is as follows:
\begin{align}\label{2.6}
    \mathbf{y}^{t-1}_j=\frac{1}{\sqrt{1-\lambda_j\beta^t}}\left(\mathbf{y}^t_j-\frac{\lambda_j\beta^t}{\sqrt{1-\gamma^{t}_j}}{\bm{\epsilon}_\theta}(\mathbf{y}^t_j,t)\right)+\sigma^t\mathbf{z},
\end{align}
with $ \gamma_j^t = \prod_{i=1}^t(1-\lambda_j\beta^i)$
for $t=T,\ldots, 0$. Note that the denoiser $\bm\epsilon_\theta$ is trained  based on categorical variables corrupted by anisotropic distributed noise.

\subsection{Imbalance-sensitive noise distribution}\label{Imbal-sen}

Let $f\in C(\mathcal F): \mathbb R^d\rightarrow \mathbb R^k$ denote a neural network, where $\mathcal F$ is the family of hypothesis class and $C(\mathcal F)$ is some proper complexity measure of the hypothesis class $\mathcal F$. We denote $L(f)=\mathbb{E}_{\mathbf{x} \sim P}[\ell(f(\mathbf{x}),\mathrm{y})]$ to be the expected error (average performance over all possible inputs), and $\widehat{L}(f)=\frac{1}{n}\sum_{i=1}^n\ell(f(\mathbf{x}_i),\mathrm{y}_i)$ to be the empirical error (performance on training data). Note that the generalization error can be regarded as an approximate estimate of the expected error. 
As discussed in \cite{hestness2017deep} and references therein, the generalization error of neural networks depends on both the model size $m$ and the dataset size $n$. Rosenfeld \emph{et al.} \cite{rosenfeld2020a} established criteria to evaluate the dependence of the error on the data size. For a given model size, scaling up the dataset results in an initial increase in performance, which then saturates to a level determined by the model size. The relationship between the expected of loss and sample size can be expressed as 
\begin{equation}\label{relation}
    L(f|m,n) \approx a(m)n^{-\alpha(m)}+b(m),
\end{equation}
where $a$, $\alpha$ and $b$ may depend on the model size $m$. 
\begin{lemma}
For $f\in C(\mathcal F)$ of the fixed size $m$, let the size of dataset $\mathcal S_1$ and $\mathcal S_2$ be $n_1$ and $n_2$ with $n_1<n_2$. Then the expected of loss $L(f|m,n_1) \geq L(f|m,n_2)$.
\end{lemma}
\begin{proof}
When we fix the model size $m$, the expected of loss \eqref{relation} only depends on the sample size $n$. As $n$ grows, $L(L|m,n)$ goes to $c_n$. For $n_1<n_2$, there is $n_1^{-\alpha(m)}>n_2^{-\alpha(m)}$. Thus, we can obtain that the expected of loss $L(f|m,n_1) \geq L(f|m,n_2)$.
\end{proof}

Rademacher complexity quantifies a model's complexity by measuring its performance on random data. A high Rademacher complexity indicates that the hypothesis space has a higher capacity, which can lead to better fitting of training data but potentially larger empirical errors. It helps us understand a model's generalization ability by comparing empirical error with expected error  \cite{cao2019learning}. By quantifying this difference, Rademacher complexity provides valuable insights into a model’s complexity and its ability to generalize.  We now discuss the Rademacher complexity of classification problems for both class-balanced and class-imbalanced tasks. Suppose $S$ is splitted into two disjoint sets, i.e., $S=V\cup T$, where $V$ denotes the validation set and $T$ denotes the training set, respectively. Let $\bm\sigma = (\sigma_1, \sigma_2,\ldots, \sigma_n)\in \{\pm 1\}^n$ to be a vector such that $V =\{\mathbf{x}_i:\sigma_i =1\}$ and $T=\{\mathbf{x}_i:\sigma_i =-1\}$. Let the variables in $\sigma$ follow a uniform distribution, such as $p[\sigma_i=1]=p[\sigma_i=-1]=\frac{1}{2}$. 

\begin{proposition}\label{prop2}
Let $f\in C(\mathcal F): \mathbb R^d\rightarrow \mathbb R^k$ be a network model of the fixed size $m$. Suppose the samples can be classified into $k$ categories, such that $\mathcal S = \mathcal S_1 \cup \mathcal S_2 \cup \ldots\cup \mathcal S_k$ with $n_1, n_2,\ldots, n_k$ being the sample size of each class.  The Rademacher complexity for the labeling function class $\mathcal F$ is given as 
 \begin{align}\label{imbalancedRC1}
 \mathcal{R}_n(\mathcal{F})=
 \mathbb E_{\bm\sigma} \left[\sum_{j=1}^{k} p_j\Big[\sup_{f\in\mathcal F}\frac1{n_j}\Big(\sum_{i=1}^{n} \chi_j(\mathbf{x}_i) \sigma_i f(\mathbf{x}_i)\Big)\Big]\right],
 \end{align} 
where $p_j$ is the proportion the $j$th category in the generalization error
\[p_j= \frac{an_j^{-\alpha}+b}{\sum_{j=1}^{k}(an_j^{-\alpha}+b)},\] 
and $\chi_j(\cdot)$ is the characteristic function of the $j$th category 
\begin{equation} 
\chi_j(\mathbf{x}_i)=\begin{cases}1,&~~ \mathbf{x}_i\in \mathcal S_j,\\
0,&~~ \mathbf{x}_i\notin \mathcal S_j,
\end{cases}
\label{chi}
\end{equation}
for $j=1,\ldots,k$.
\end{proposition}

\begin{proof}
Let $L_{\mathcal{S}_j}(f)$ denote the expected loss of the $j$-th class for $j=1,\ldots, k$. According to \eqref{relation}, the contribution of $j$-th class to Rademacher complexity can be expressed as follows:
\begin{equation}\label{class-ratio}
    p_j(\mathbf x| \mathbf y=j)= \frac{an_j^{-\alpha}+b}{\sum_{j=1}^{k}(an_j^{-\alpha}+b)}.
\end{equation} 
For a given $\bm\sigma$, the Rademacher complexity of $f$ can be estimated by 
\begin{equation}\label{rc}
    \begin{split}
    \mathcal{R}_n(\mathcal{F})&=\sup_{f\in\mathcal F} \Big(L_{V}(f) - L_{T} (f)\Big)\\
    &=\sum_{j=1}^k  p_j \Big[\sup_{f\in\mathcal F} \Big(L_{V\cap S_j}(f) - L_{T\cap S_j} (f)\Big)\Big] \\
    &=\sum_{j=1}^k  p_j \Big[  
 \sup_{f\in\mathcal F} \frac{1}{n_j} \Big(\sum_{i=1}^{n} \chi_j(\mathbf x_i) \sigma_i f(\mathbf x_i)\Big) \Big]. 
    \end{split}
\end{equation}
Let the variables in $\bm\sigma$ be distributed i.i.d. according to $p[\sigma_i=1]=p[\sigma_i=-1]=\frac{1}{2}$. The Rademacher complexity of $\mathcal F$ with respect to $S$ can be obtained as follows:
\begin{equation*}\label{imbalancedRC2}
 \mathcal{R}_n(\mathcal{F})=\mathbb E_{\bm\sigma}\left[\sum_{j=1}^{k} p_j \Big[\sup_{f\in\mathcal F}\frac1{n_j}\Big(\sum_{i=1}^{n} \chi_j(\mathbf{x}_i) \sigma_i f(\mathbf{x}_i)\Big)\Big]\right],
 \end{equation*} 
 which completes the proof.
\end{proof}
As a special case, we have $p_1 = p_2= \ldots = p_k = \frac1k$ and $n_1 = n_2= \ldots = n_k = \frac Nk$ for the class-balanced classification problem. Therefore, we can obtain the following balanced Rademacher complexity as follows:
\[\mathcal{R}_n(\mathcal{F})=\mathbb E_{\bm\sigma}\left[\sum_{j=1}^{k} p_j \Big[\sup_{f\in\mathcal F}\frac1{n_j}\Big(\sum_{i=1}^{n} \chi_j(\mathbf{x}_i) \sigma_i f(\mathbf{x}_i)\Big)\Big]\right]
        =\frac{1}{n} \mathbb E_{\bm\sigma}\Big[\sup_{f\in\mathcal F} \sum_{i=1}^{n} \sigma_i f(\mathbf{x}_i)\Big].\]

 Bartlett and Mendelson \cite{bartlett2002rademacher} provided a generalization bound in terms of the Rademacher complexity, which can be given in term of imbalanced Rademacher complexity. 
\begin{theorem}\label{them} 
Assume the loss $\ell$ is Lipschtitz with respect to the first argument and $\ell$ is bounded by $c$. For any $\delta >0$ and with probability at least $1-\delta$ over the randomness of the training data, all hypotheses $f \in C(\mathcal{F})$ have the following the generalized loss
\begin{equation}\label{BM}
    L(f) \leq \sum_{j=1}^{k} p_j\left(\widehat{L}_j(f)+2L_{\ell}\mathcal{R}_{n_j}(\mathcal{F})+c\sqrt{\frac{\log(1/\delta)}{2n_j}}\right),
\end{equation}
where $\widehat{L}_j(f)$ and $\mathcal{R}_{n_j}(\mathcal{F})$ denotes the empirical error and Rademacher complexity of the $j$-th class, respectively. 
\end{theorem}
\begin{proof}
Similar to the generalization error \eqref{rc} in the class imbalance problem, the empirical error of the $j$-th class can be defined as
\[\widehat{L}_j(f) = \frac{1}{n_j}\sum_{i=1}^{n}\chi_j(\mathbf{x}_i)\ell(f(\mathbf{x}_i),\mathrm{y}_i).\]
Then, the generalization bound for $j$-th class is the following inequality according to \cite{bartlett2002rademacher}: 
\begin{equation}\label{BM1}
Lj(f) \leq \widehat{L}_j(f)+2L_{\ell}\mathcal{R}_{n_j}(\mathcal{F})+c\sqrt{\frac{\log(1/\delta)}{2n_j}}.
\end{equation}
Based on the generalization loss for each individual class, we then apply a union bound across all classes to obtain 
\[L(f) = \sum_{j=1}^{k} p_j L_j(f) \leq \sum_{j=1}^{k} p_j\left(\widehat{L}_j(f)+2L_{\ell}\mathcal{R}_{n_j}(\mathcal{F})+c\sqrt{\frac{\log(1/\delta)}{2n_j}}\right),\]
which completes the proof. 
\end{proof}
Theorem \ref{them} indicates that improving the generalization of minority classes is important for controlling the generalization error of the classification function. The proportion of generalization error contributed by different classes can be used to guide the diffusion speed in the process. Therefore, we determine the noise intensity of each category in  forward process based on $p_j$ to accelerate the diffusion of tail classes. 
As shown in Figure \ref{motivation2}, we adopt noise level to each class according to its contribution to the generalization error to achieve a balance between head classes and tail classes. More specifically, the diffusion speed is determined by $\bm\lambda$ in \eqref{e10} for our anisotropic diffusion process, which should be inbetween $[1,c\nu]$ with $\nu$ denoting the imbalance ratio\footnote{$\nu=\max_{k}\left\{n_k\right\} /\min_{k}\left\{n_k\right\}$ 
with $n_k$ being the numbers of samples in the $k$-th class.} and $c$ is a positive constant to control the growth rate of noise level among different classes. Therefore, we define $\lambda_j$ as follows:
\begin{equation}\label{lambda}
\lambda_j = c\nu \Big(\frac{n_j^{-\alpha}}{\sum_{j=1}^kn_j^{-\alpha}}\Big) + 1, \quad \mbox{for} ~j = 1,\ldots,k,
\end{equation}
where $\alpha$ and $c$ are estimated by fitting the dataset. More details on implementation qw21sa  can be found in numerical experiments.



\section{Priors for Diffusion Probabilistic Model}\label{priors}

\begin{figure*}[t]
\centering
    \begin{subfigure}{\includegraphics[scale=0.3]{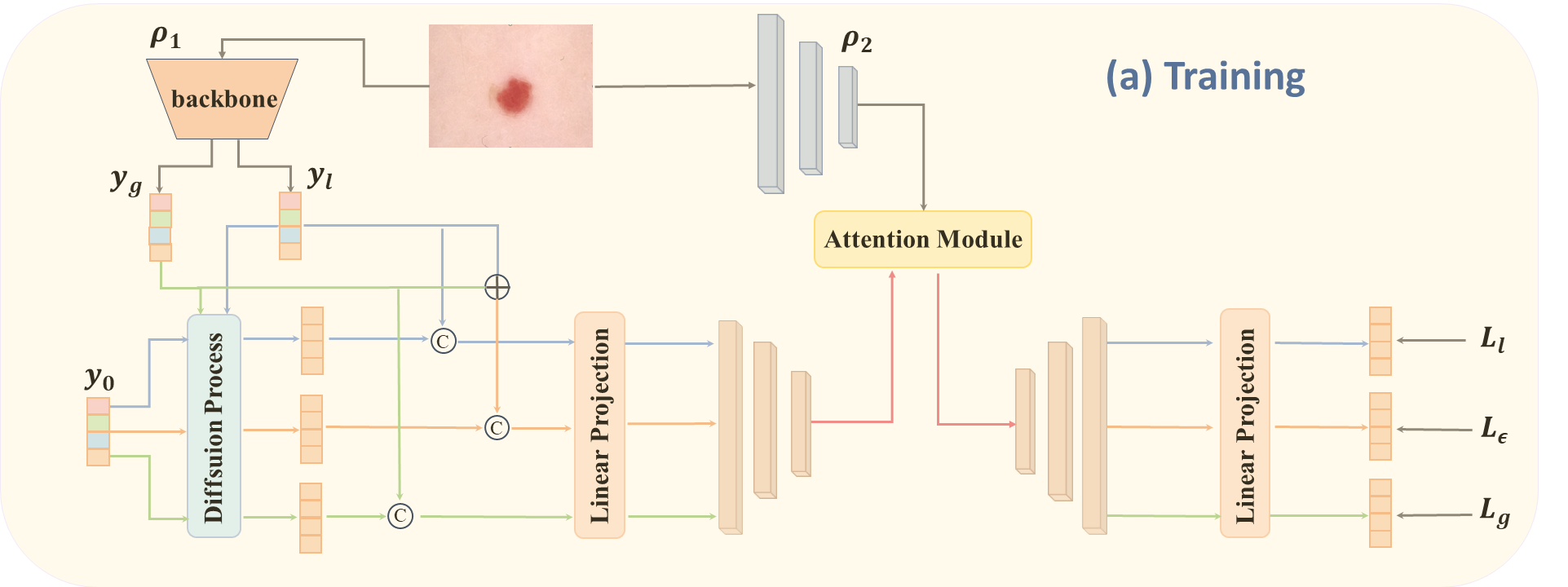}}
    \end{subfigure}
    \begin{subfigure}
        {\includegraphics[scale=0.3]{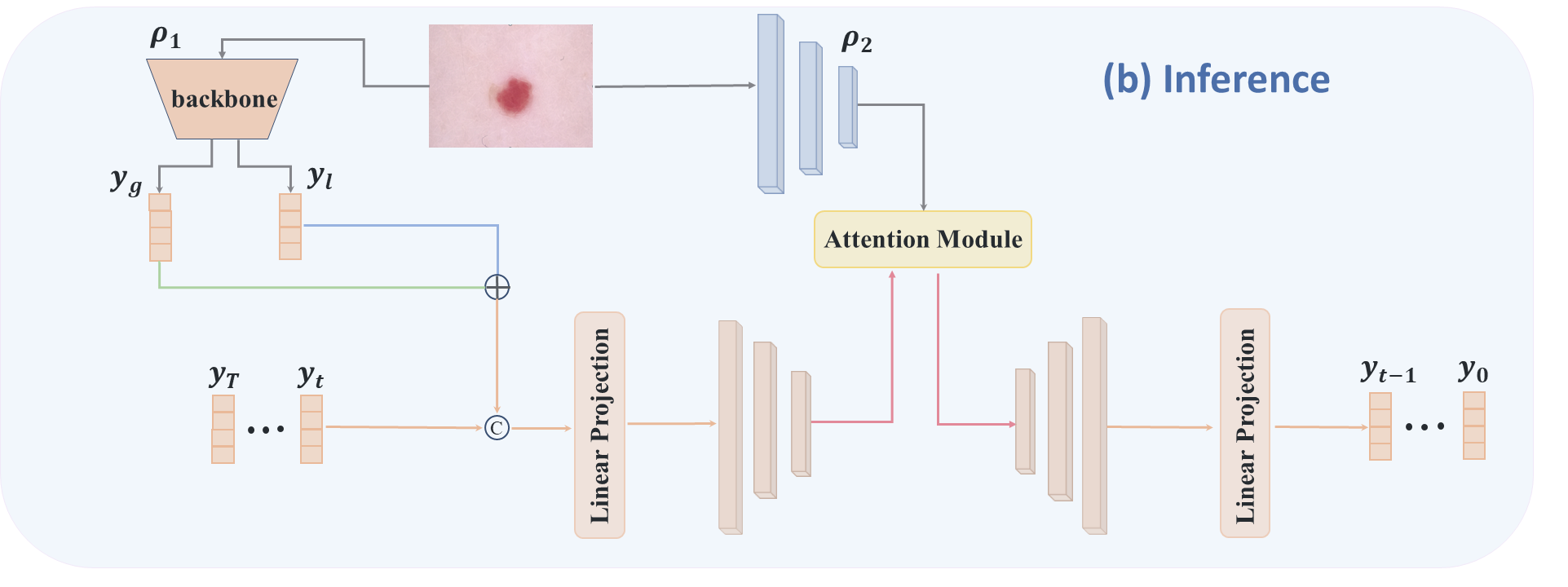}}
    \end{subfigure}
    \caption{Overview of our anisotropic diffusion probabilistic model, where (a) the training phase (forward process) and (b) the inference phase (reverse process). }
    \label{model}
\end{figure*}

\subsection{Forward prior in diffusion model}\label{forwardprior}
Effective prior knowledge is crucial for the success of image classification tasks. It not only enhances the performance of models but also improves their interpretability and adaptability, making the models more promising and practical for real-world applications. Dermatologists typically base their diagnoses on prior knowledge of the shape, color, size of skin lesions, and the patient's medical history. Zhou \emph{et al.} \cite{zhou2023novel} proposed a multi-task model by imitating the diagnostic procedures of dermatologists, which is capable of simultaneously predicting the body parts, lesion attributes, and the disease itself. By effectively utilizing prior knowledge about disease diagnosis, the accuracy, interpretability, and reliability of the diagnostic model is obviously improved. Similarly, Yang \emph{et al.} \cite{yang2023diffmic} utilized both global and local image features to obtain a fual guidance diffusion network for medical image classification. Both works validate that global priors are effective in capturing long-range dependencies and contextual information, while local priors excel at managing local consistency and relationships, thereby modeling texture and attribute information.

Conditional priors also play important regulating roles in diffusion probability models, which can guide the noise generation process in each diffusion step to improve the quality and authenticity of generated data. Suppose additional information is introduced as conditions for $p(\mathbf{y}|\mathbf{x})$, which allows to explicit control over the generated data by incorporating conditioning information $\mathbf{x}$. Then the reverse process \eqref{e3} can be reformulated as follows:
\begin{equation}\label{e6}
    p\left(\boldsymbol{\mathbf{y}}^{0: T} \mid \mathbf{x}\right)=p\left(\boldsymbol{\mathbf{y}}^{T}\right) \prod_{t=1}^{T} p_{\boldsymbol{\theta}}\left(\boldsymbol{\mathbf{y}}^{t-1} \mid \boldsymbol{\mathbf{y}}^{t}, \rho(\mathbf{x})\right),
\end{equation}
where $\rho(\mathbf{x})$ can be selected depending on the target task. 

More specifically, we incorporate both global and local dynamic feature priors into the forward diffusion process. The prior network $\rho_1$ is used to extract both global and local feature from original images. For the global branch, the input image undergoes global feature encoding, followed by a $1\times1$ convolution to obtain an attention map, where the average image of the attention map is the final global prior. Additionally, regions with higher attention are cropped as inputs for the local branch, from which local priors are extracted and fused to form the final local features. Finally, the global and local priors are integrated to enhance the features and structures of the data in each diffusion step; see Figure \ref{model} for more details.  Suppose the global and local prior are $\mathbf{y}_g \in \mathbb{R} ^ {k}$ and $\mathbf{y}_l \in \mathbb{R} ^ {k}$, and noisy global and local variables are $\mathbf{y}^{t}_g $ and $\mathbf{y}^{t}_l$. For simplicity, let $\widehat{\mathbf{y}} = \{\mathbf{y}_g,\mathbf{y}_f,\mathbf{y}_l\}$ and $\widehat{\mathbf{y}}^t:=\{\mathbf{y}^t_g,\mathbf{y}^t_f,\mathbf{y}^t_l\}$, where $\mathbf{y}_f = \frac{1}{2}(\mathbf{y}_g+\mathbf{y}_l)$.
Then we have 
\begin{equation}\label{e7}
\begin{aligned}
    \widehat{\mathbf{y}}_j^t &= \sqrt{\gamma_j^t}\mathbf{y}^0+ \sqrt{1-\gamma_j^t} \bm \epsilon+\big (1-\sqrt{\gamma_j^t} \big)\widehat{\mathbf{y}}.
\end{aligned}
\end{equation}
We then concatenate the noisy variable $\widehat{\mathbf{y}}^t$ with the features $\widehat{\mathbf{y}}$, which are further passed through a linear projection layer prepared for network training.

\subsection{Reverse prior in diffusion probabilistic model}\label{reverseprior}

In the classical DDPM approach, the noisy variable $\mathbf{y}^{t}$ is the sole input for the denoising network, used to characterize data distribution information. According to model \eqref{e5},  sampling from noisy variables may introduce inherent randomness during inference, resulting in different outcomes with each sampling execution. Therefore, it is necessary to incorporate effective prior information from original images to improve classification accuracy. Hence, we propose the feature-conditioned embedding as shown in Figure \ref{model}. Specifically, we feed the original image $\mathbf{x}$ into a ResNet encoder $\rho_2(\cdot)$ and process the output through a cross-attention module as follows:
\begin{align*}
Q=\rho_2(\mathbf{x})W_{Q}, \quad K=E\big(f\left([\widehat{\mathbf{y}}^{t}, \widehat{\mathbf{y}}]\right)\big)W_{K}, \quad \mbox{and}~~
V=E\big(f(\left[\widehat{\mathbf{y}}^{t}, \widehat{\mathbf{y}}\right])\big)W_{V}, 
\end{align*}
where $f(\cdot)$ represents a linear projection layer, $E(\cdot)$ denotes the encoder of the denoising network, and $W_{Q}, W_{K},W_{V}$ represents the parameter matrices that the network needs to be learnt. Thus, we have
\begin{equation*}\label{e12}
\mbox{Attention}(Q,K,V) =\mbox{Softmax}\Big(\frac{QK^{T}}{\sqrt{d}}\Big)V,
\end{equation*}
where $d$ is the feature dimension. We then integrate the cross-attention module into the denoising model to embed lesion image features, and enable accurate learning of the classification mapping and focusing the model on high-level semantics, thereby obtaining more robust feature representations.

Additionally, prior information is incorporated into both the diffusion process and the denoising network to focus the model more on high-level semantic features, facilitating precise learning of the classification mapping, defined by
\begin{equation*}\label{e13}
\epsilon_{\theta}\left(\rho_2(\mathbf{x}), \widehat{\mathbf{y}}^{t}, \widehat{\mathbf{y}}, t\right)=D\Big(\operatorname{Att}\big(\rho_2(\mathbf{x}), E\big(f\left(\left[\widehat{\mathbf{y}}^{t}, \widehat{\mathbf{y}}\right]\big)\big), t\right), t\Big),
\end{equation*}
where $D(\cdot)$ denotes the decoder of the denoising network. The noisy variable $\widehat{\mathbf{y}}^{t}$ is sampled in the diffusion process based on the prior $\widehat{\mathbf{y}}$ and $\operatorname{Att}(\cdot)$ is the cross attention module used to fuse image features in the denoising model. Note that the time step $t$ is encoded into the network, enabling the model to accurately identify the magnitude of the noise during both training and inference stage.

\subsection{Training objective and algorithm}\label{objective}

We adopt a multi-task loss function to simultaneously learn global features, local features, and classification results. Let $\epsilon^f$ and $\epsilon^g$ denote the global feature and local feature obtained from the denoising network, respectively. Similar to DiffMIC \cite{yang2023diffmic}, we use the Maximum Mean Discrepancy (MMD) regularization loss to learn the similarity between the sampled noise and the estimated Gaussian noise. The MMD is widely used in transfer learning, particularly in domain adaptation, which measures the distributional distance between two different but related random variables using kernel functions. 
Specifically, for the global branch, the MMD loss is defined as 
\begin{align*}
\mathcal{L}_{g}(\epsilon \| \epsilon_g) = \mathcal{K}\left(\epsilon, \epsilon\right)-2 \mathcal{K}(\epsilon_g, \epsilon)+\mathcal{K}\left(\epsilon_g, \epsilon_g\right),
\end{align*}
where $\epsilon_g = \epsilon_{\theta}\left(\rho_2(\mathbf{x}),\mathbf{y}^t_g, \mathbf{y}_g, t\right)$ and $\mathcal{K}(\cdot,\cdot)$ is a positive definite kernel within a reproducing Hilbert space. Similarly, we define the MMD loss for the local branch as 
\begin{align*}\label{e20}
    \mathcal{L}_{l}(\epsilon \| \epsilon_l) = \mathcal{K}\left(\epsilon, \epsilon\right)-2 \mathcal{K}(\epsilon_l, \epsilon)+\mathcal{K}\left(\epsilon_l, \epsilon_l\right),
\end{align*}
where $ \epsilon^l = \epsilon_{\theta}\left(\rho_2(\mathbf{x}), \mathbf{y}^t_l, \mathbf{y}_l, t\right)$.
For classification task, the loss is consist with DDPM defined following the variational bound as follows:
\begin{equation*}\label{e21}
    \mathcal{L}_{\epsilon}=\|\epsilon-\epsilon_f\|^{2}, \quad 
    \text { with } \epsilon_f=\epsilon_{\theta}\left(\rho_2(\mathbf{x}),\mathbf{y}^{t}_{f},\mathbf{y}_{f}, t\right).
\end{equation*}
Therefore, the total loss $\mathcal{L}_{total}$ for the denoising network is:
\begin{equation}\label{e22}
    \mathcal{L}_{total}= w(\mathcal{L}_{g}+\mathcal{L}_{l})+\mathcal{L}_{\epsilon},
\end{equation}
where $w$ is a positive hyper-parameter.  We further specify our sampling algorithm for utilizing the pre-trained ADPM for image classification tasks. We assume both training and sampling dataset follow the same distribution. During sampling, the imbalance-sensitive noise level  is estimated by the classification results of the prior extraction model $\rho_1$ as follows:
\begin{equation}\label{inference}
 \lambda = c\Lambda \Big(\frac{n_k^{-\alpha}}{\sum_{j=1}^k n_j^{-\alpha}}\Big) +1,~~ k = \arg\max_{j=1,\ldots,k}\mbox{softmax}\big(\rho_1(\mathbf x)\big),
\end{equation}
which is used to obtain anisotropic noise distribution. 
To sum up, we conclude the detailed anisotropic sampling algorithm as Algorithm \ref{alg:algorithm1}.

\begin{algorithm}[htb]  
\caption{Anisotropic sampling method}  
\label{alg:algorithm1}
  \begin{algorithmic}[1]     
    \STATE Define original image $\mathbf{x}$, and sequence $\{\beta^t\}_{t=1}^{T}$
    \STATE Compute $\lambda$ according to Eq. \eqref{inference};
    \STATE Compute $\bm\gamma$ according to Eq. \eqref{2.6};
    \STATE $\mathbf{y}^T \sim \mathcal{N}(\mathbf{y}_f,\mathbf{I})$; 
    \STATE \textbf{for} $t=T$ to $1$ \textbf{do}
    \STATE \quad $\mathbf{z} \sim \mathcal{N}
    (\mathbf{0},\mathbf{I})$;
    \STATE \quad $\sigma^t=\sqrt{\frac{\lambda\beta^t\big(1-\gamma^{t-1}\big)}{1-\gamma^t}}$;
    \STATE \quad $
        \mathbf{y}^{t-1}=\frac{1}{\zeta^t}\left(\mathbf{y}^t-\frac{\xi^t-\zeta^t}{\xi^t}{\mathbf{y}_f} -\frac{\lambda\beta^t}{\sqrt{\xi^t}}{\bm{\epsilon}_\theta}(\mathbf{y}^t,t)\right)+\sigma^t\mathbf{z}
        $ with $\zeta^t=\sqrt{1-\lambda\beta^t}$ and $\xi^t = 1-\gamma^t$;
    \STATE \textbf{end for}
    \STATE \textbf{Return}  $\mathbf y^0$
  \end{algorithmic}  
\end{algorithm}

\section{Numerical experiments}\label{experiments}

\subsection{Implementation details}\label{details}

Our model is implemented in PyTorch and executed on a NVIDIA GeForce RTX 3090 GPU. The prior network can adopt various network model, such as the ResNet \cite{yang2023diffmic}, ViT \cite{zhou2023novel} and Swin transformer \cite{liu2021swin} etc., where we used ResNet as the default prior network in our work. By using the anisotropic noise in the forward diffusion process, the noisy variables are concatenated with the feature priors and then fed into a linear layer with an output dimension of 6144, resulting in the fused vector in the latent space. The denoising network consists of linear layers and transformer modules to leverage the transformer to embed image priors. 
The training process of our anisotropic diffusion model is similar to the traditional DDPM \cite{ho2020denoising}. The step size $t$ is uniformly selected from the range $[1,T]$, and the noisy variance $\beta$ is linearly selected with $\beta_1=0.0001$ and $\beta_T=0.02$. During the overall training process, the prior network has a pre-warmup phase of 15 epochs before being jointly trained with the diffusion model. We employ center cropping to resize the images to $224\times224$, and apply random flipping and rotation for data augmentation. The network is optimized by the Adam optimizer by training over 1000 epochs with a batch size of 32. The initial learning rate for the denoising network is 0.001. We gradually increase the learning rate at the beginning of training helps stabilize the process, while later, the learning rate decays following a half-cycle cosine function. Throughout the training stage, the total step size for the forward process of the diffusion model is maintained at 1000 steps, while during the testing stage, the total step size is empirically determined to be 250.
When calculating the total loss, we choose the radial basis function (RBF) as the kernel, and $w$ is empirically set as 0.5.

\subsection{Datasets}\label{Datasets}
The effectiveness of our network model was evaluated on three public datasets used for medical image classification, which are  PAD-UFES \cite{pacheco2020impact}, HAM10000 \cite{tschandl2018ham10000}, SCIN \cite{ward2024crowdsourcing}, and Hyper-Kvasir \cite{Borgli2020}. (1) \textbf{PAD-UFES:} The dataset contains 2298 clinical images of 6 types of skin diseases. Each sample comprises a clinical image accompanied by metadata that includes labels for diseases and body parts. (2) \textbf{HAM10000:} The dataset is derived from the challenge hosted by The International Skin Imaging Collaboration in 2018, aimed at classifying skin diseases. It comprises of 10,015 dermatoscopic images, totaling seven types of skin diseases. (3) \textbf{SCIN:} Different from the data utilized by the Google Health team in their dermatological diagnostic system \cite{liu2020deep}, the SCIN dataset\footnote{https://github.com/google-research-datasets/scin}, collected from U.S. Google Search users through voluntary image donations, contains over 5,000 contributions of common dermatology conditions. We exclude cases with poor image quality and no diagnoses, resulting in a final dataset of 3061 cases. (4) \textbf{Hyper-Kvasir:} Hyper-Kvasir is the largest image and video dataset of the gastrointestinal tract available, which are categorized into 23 classes. 

\begin{table}[t]
\renewcommand{\arraystretch}{1.5}
\centering
\caption{The summary of the imbalance ratio for the datasets used in our experiments.}\label{IR} 
\begin{tabular}{cccccc}
\toprule
Dataset & Classes  & Sample size & Maximal size & Minimal size& Imbalance ratio \\
    \hline
    PAD-UFES & 6 & 2298 & 845 & 52 & 16 \\ 
    HAM10000 & 7 & 10015 & 6705 & 115 & 58 \\ 
    SCIN  & 20 & 2744 & 1078 & 3 & 359 \\  
    Hyper-Kvasir & 23 & 10662 & 1148 & 6 & 191 \\
\bottomrule
    \end{tabular}
\end{table}

We selected data from different modalities for testing, including clinical skin images, dermatoscopic images, and endoscopic images, to demonstrate the generalizability of our proposed classification model. As shown in Table \ref{IR}, these datasets represent typical class imbalance classification problems, which can be measured by the Imbalance Ratio (IR) to assess the degree of imbalance in the datasets. Note that if the IR approaches 1, it indicates that the data is relatively balanced; the larger the IR value, the more severe the data imbalance. Additionally, the difficulty of classification is also affected by the amount of data. Although the imbalance ratio of HAM10000 is larger than that of PAD-UFES, the sample size of PAD-UFES dataset is much smaller than that of HAM10000, which also makes its classification challenging.  
Similar to \cite{gong2020distractor}, we divided HAM10000 into a training set and a test set with a 7:3 ratio. For PAD-UFES, SCIN and Hyper-Kvasir dataset, we conducted experiments using five-fold cross-validation.

\subsection{Evaluation on imbalance-sensitive noise}\label{eva}
On the first place, we discuss the choices of imbalance-sensitive noise to the performance of our ADPM on classification accuracy. In particular, we vary both the tunable variable $\alpha$ and $c$ in \eqref{lambda} and evaluate the classification performance on PAD-UFES dataset. In Table \ref{tab2}, we exihibit the F1-score w.r.t. different combinations of $\alpha$ and $c$, where $\alpha$ is chosen inbetween $[0,1]$ and $c$ is chosen inbetween $[1,10]$. As can be observed, F1-score varies with the choices of $\alpha$ and $c$. Specifically, when $\alpha=0$, our ADPM reduces to original DDPM, and the F1 score does not change with variations in $c$. And the best F1-score is achieved by setting $\alpha=1/6$ and $c=5$, which gives a more than 4\% improvement in F1-score. The results demonstrate the anisotropic noise can enhance inter-class diversity and increase the attention to tail classes. Since both $\alpha$ and $c$ are assumed to depend on the model size $m$ in \eqref{relation}, i.e., when the model size remains the same, $\alpha$ and $c$ can be fixed across different datasets, we choose $\alpha = 1/6$ and $c = 5$ for all datasets.

\begin{table}[h]
\caption{F1-scores obtained by our ADPM with different combinations of the variable $\alpha$ and $c$ on the PAD-UFES dataset.}\label{tab2}
\centering
\setlength{\tabcolsep}{7pt}
\begin{tabular}{c|c|c|c|c|c|c|c|c|c|c}
\toprule
\diagbox{$\alpha$}{c} & 1 & 2 & 3 & 4 & 5 & 6 & 7 & 8 & 9 & 10 \\
\hline
0 & 0.678 & 0.678 & 0.678 & 0.678 & 0.678 & 0.678 & 0.678 & 0.678 & 0.678 & 0.678  \rule{0pt}{3ex} \\ 
\hline
1/6 & 0.682 & 0.683 & 0.681 & 0.671 & \textbf{0.721} & 0.682 & 0.698 & 0.673 & 0.687 & 0.669 \rule{0pt}{3ex} \\
\hline
1/4 & 0.651 & 0.673 & 0.668 & 0.675 & 0.693 & 0.675 & 0.673 & 0.679 & 0.672 & 0.646 \rule{0pt}{3ex} \\
\hline
1/2 & 0.677 & 0.655 & 0.661 & 0.654 & 0.668 & 0.667 & 0.669 & 0.646 & 0.645 & 0.662 \rule{0pt}{3ex} \\
\hline
1 & 0.669 & 0.653 & 0.631 & 0.638 & 0.661 & 0.667 & 0.651 & 0.664 & 0.654 & 0.652 \rule{0pt}{3ex} \\

\bottomrule
\end{tabular}
\end{table}

\begin{figure*}[t]
\centering
    \subfigure[$\alpha=0$]{
        \centering
        \includegraphics[width=0.197\textwidth]{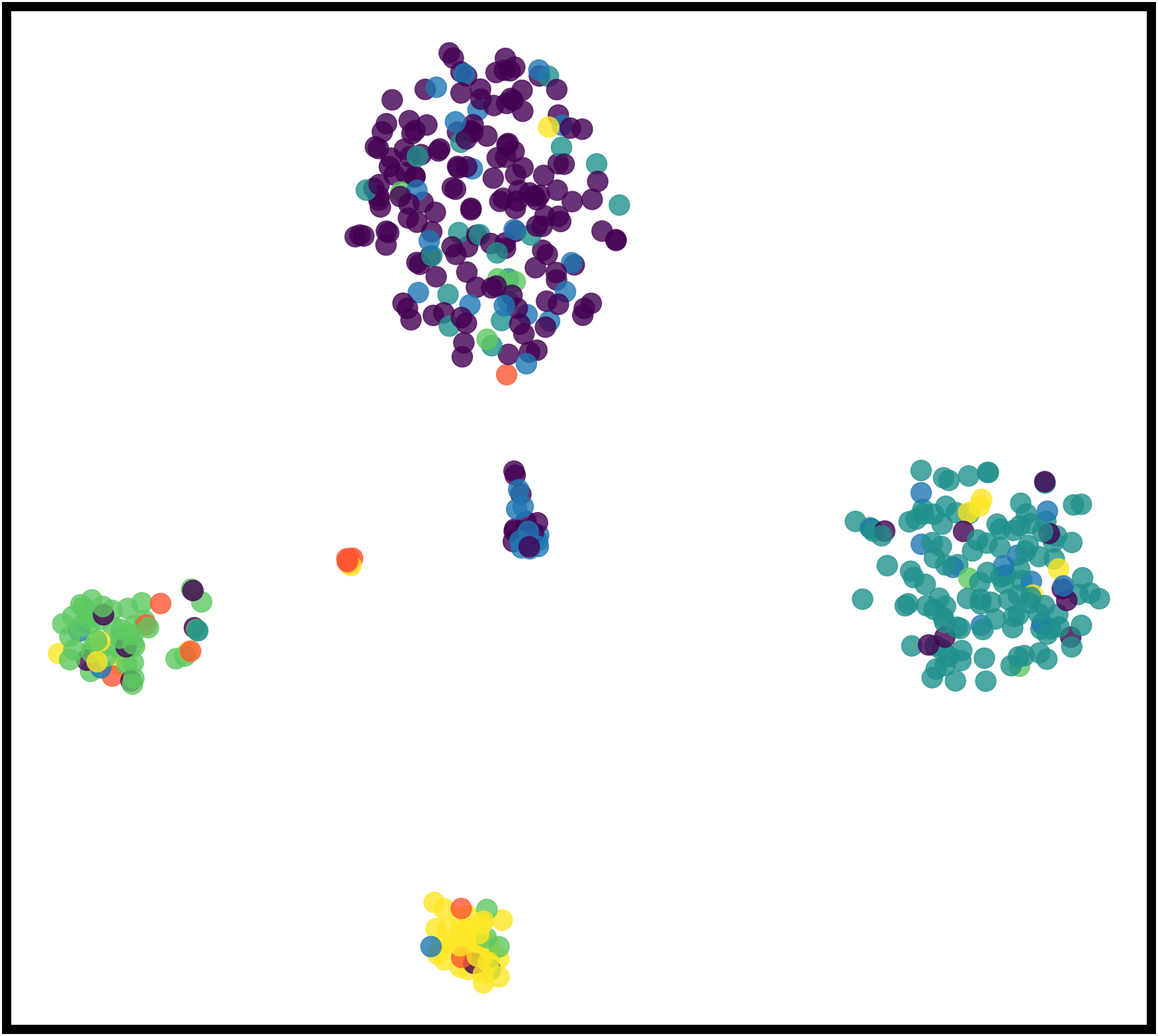}
    }\hspace{-2.2ex}
    \subfigure[$\alpha=1/6$]{
       \centering
        \includegraphics[width=0.197\textwidth]{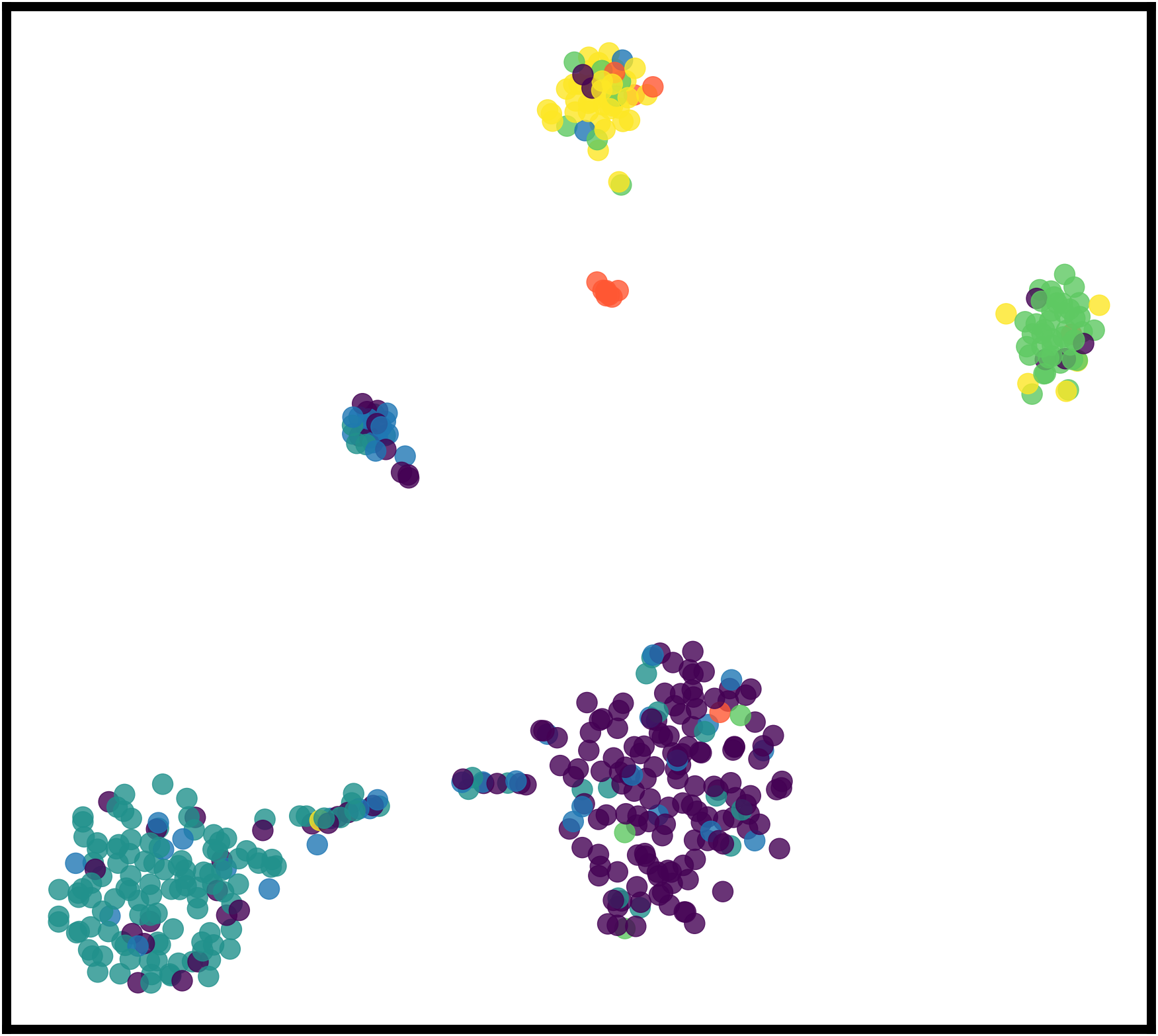}
    }\hspace{-2.2ex}
    \subfigure[$\alpha=1/4$]{
        \centering
        \includegraphics[width=0.197\textwidth]{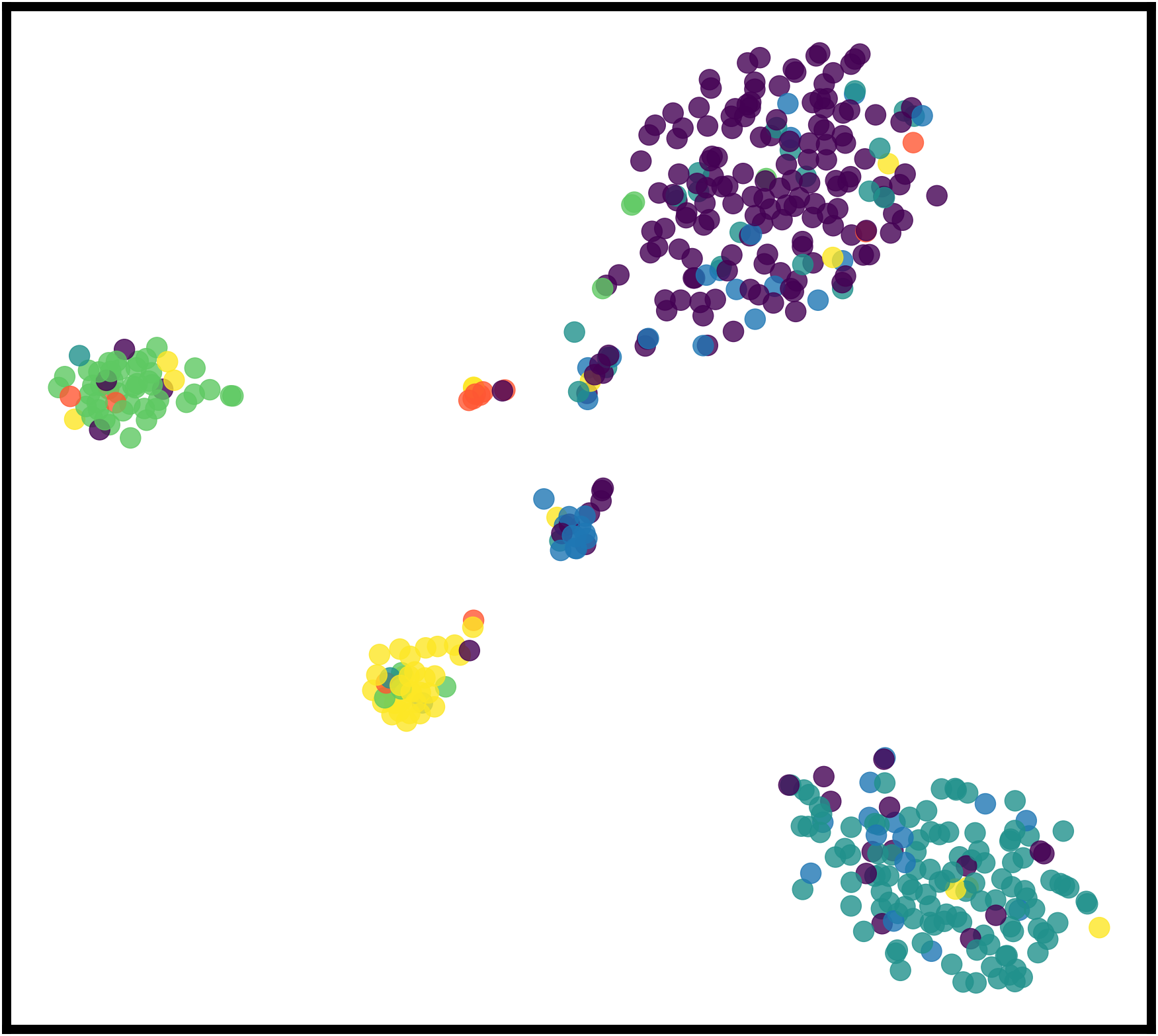}
    }\hspace{-2.2ex}
    \subfigure[$\alpha=1/2$]{
        \centering
        \includegraphics[width=0.197\textwidth]{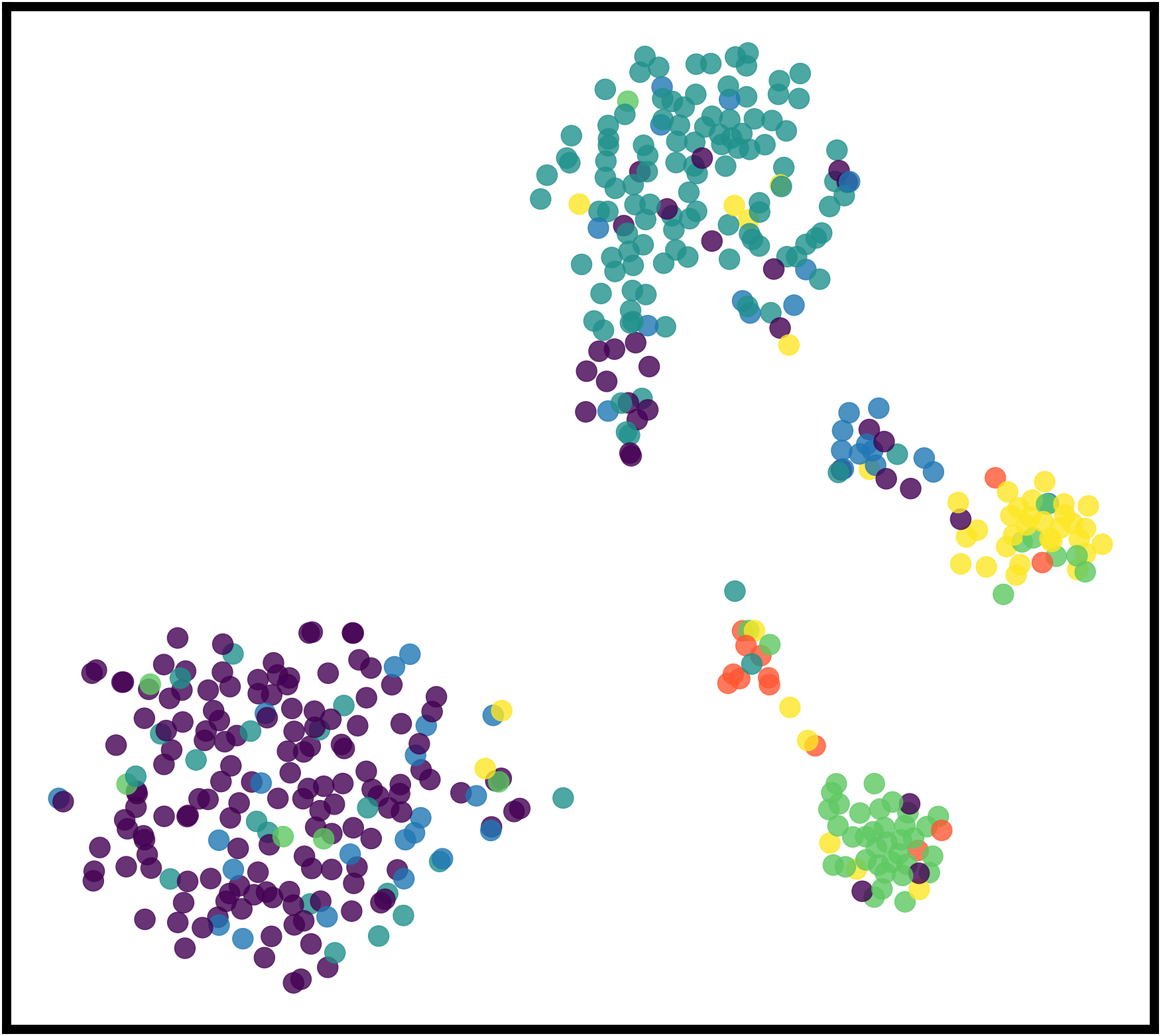}
    }\hspace{-2.2ex}
    \subfigure[$\alpha=1$]{
        \centering
        \includegraphics[width=0.197\textwidth]{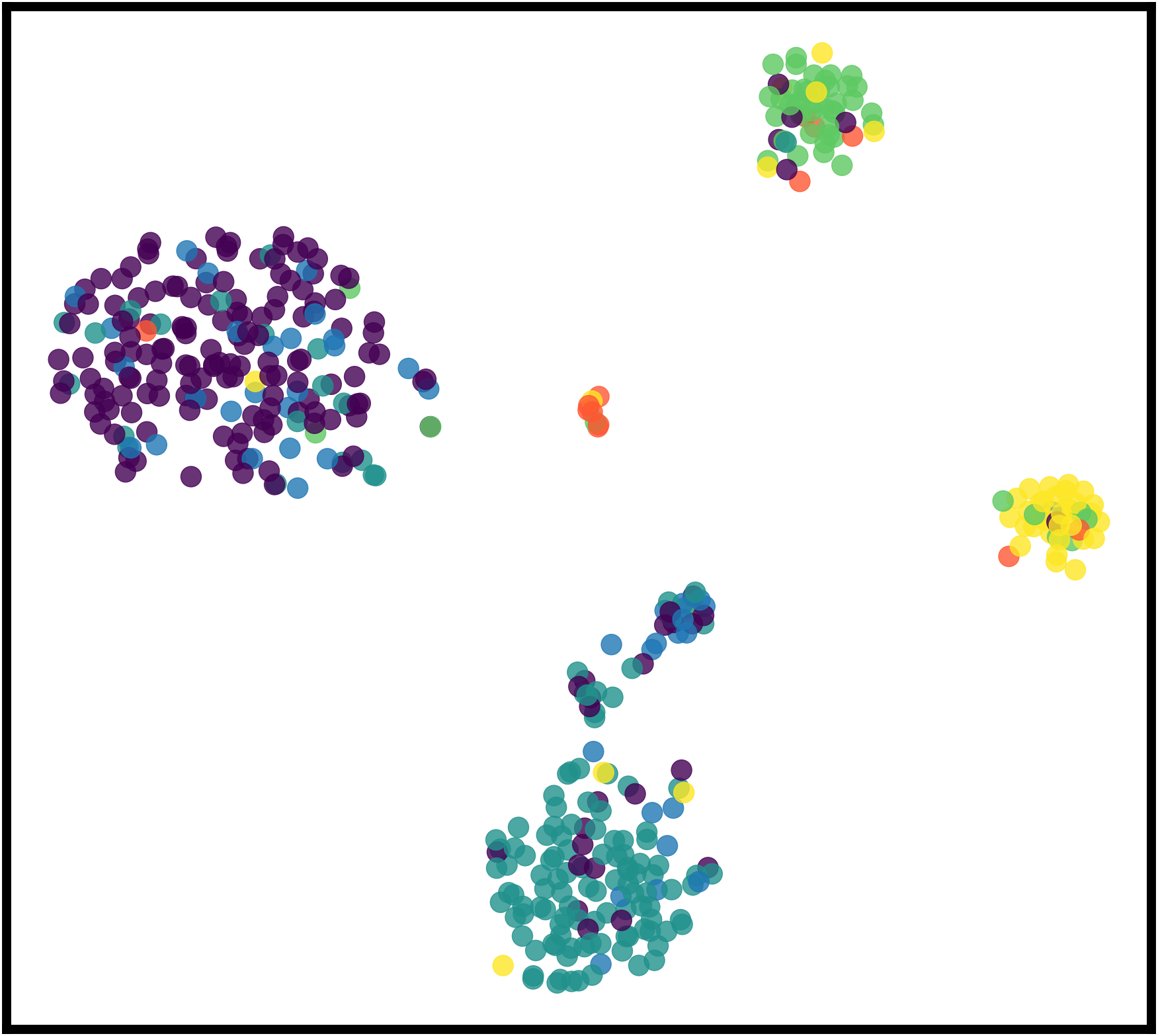}
    }\hspace{-2.2ex}
\caption{The t-SNE obtained from the final feature embedding by the diffusion reverse
process on PAD-UFES dataset, where $c$ is chosen as $c=5$ and $\alpha$ is chosen from $[0,1]$. \label{t-sne}}
\label{noise}
\end{figure*}

Besides, we used the t-SNE tool to visualize the denoised feature embeding at time $t=0$ to compare the classification accuracy. As shown in Figure \ref{noise}, the distribution of predicted classes on the PAD-UFES dataset varies as we adopted different values of $\alpha$ and fixed $c=5$ in our ADPM. When $\alpha=1/6$, the sample distribution within the same class is more concentrated than others, providing the more accurate classification results. 

\subsection{Ablation study}\label{ablationstudy}
We conducted extensive experiments on the PAD-UFES dataset to validate the effectiveness of the conditional embedding module in both forward and backward diffusion process. We designed three network models for comparison, and the results are shown in Table \ref{Ablation}. The baseline network model discarded the diffusion process, which is the ResNet network for image classification. Subsequently, we introduced the anisotropic diffusion model and incorporated the conditional embeddings in the forward diffusion process, denoted as ``V1''. Alternatively, we introduced the anisotropic diffusion model and incorporated image feature fusion in the reverse denoising model, denoted as ``V2''. The last model is our proposed one integrating the anisotropic diffusion with both prior modules. As we can see in Table \ref{Ablation}, both ``V1'' and ``V2'' models provide better accuracy and F1-score than the baseline model, and the effect of the reverse prior is more important than the forward prior improving the model performance. As shown, our method achieves the best accuracy and F1-score, indicating that both the forward prior and the reverse prior contribute to improved classification performance. 
From the comparison, it is evident that the combination of anisotropic diffusion and prior modules significantly improves the accuracy and F1-score, thereby validating the effectiveness and superiority of the proposed classification model.

\begin{table}[h]
\centering
\caption{Ablation studies on our ADPM w/ and w/o anisotropic noise, forward and reverse prior on the PAD-UFES dataset.}\label{Ablation} 
\begin{tabular}{p{1.5cm}|c|c|c|c|c}
\toprule
Model     & Anisotropic noise  & Forward prior & Reverse prior& Accuracy & F1-score\\
    \hline
    Baseline &  &  &    & 0.623 & 0.511 \\
    
    V1 & \Checkmark & \Checkmark &  & 0.745 & 0.663 \\
    
    V2 & \Checkmark &  & \Checkmark & 0.768 & 0.676\\
    \hline
    \textbf{ADPM} & \Checkmark & \Checkmark & \Checkmark & \textbf{0.772} & \textbf{0.721} \\
 \bottomrule
    \end{tabular}
\end{table}

\subsection{Numerical convergence}
\begin{figure*}[h]
\centering
\subfigure{\includegraphics[width=0.165\textwidth]{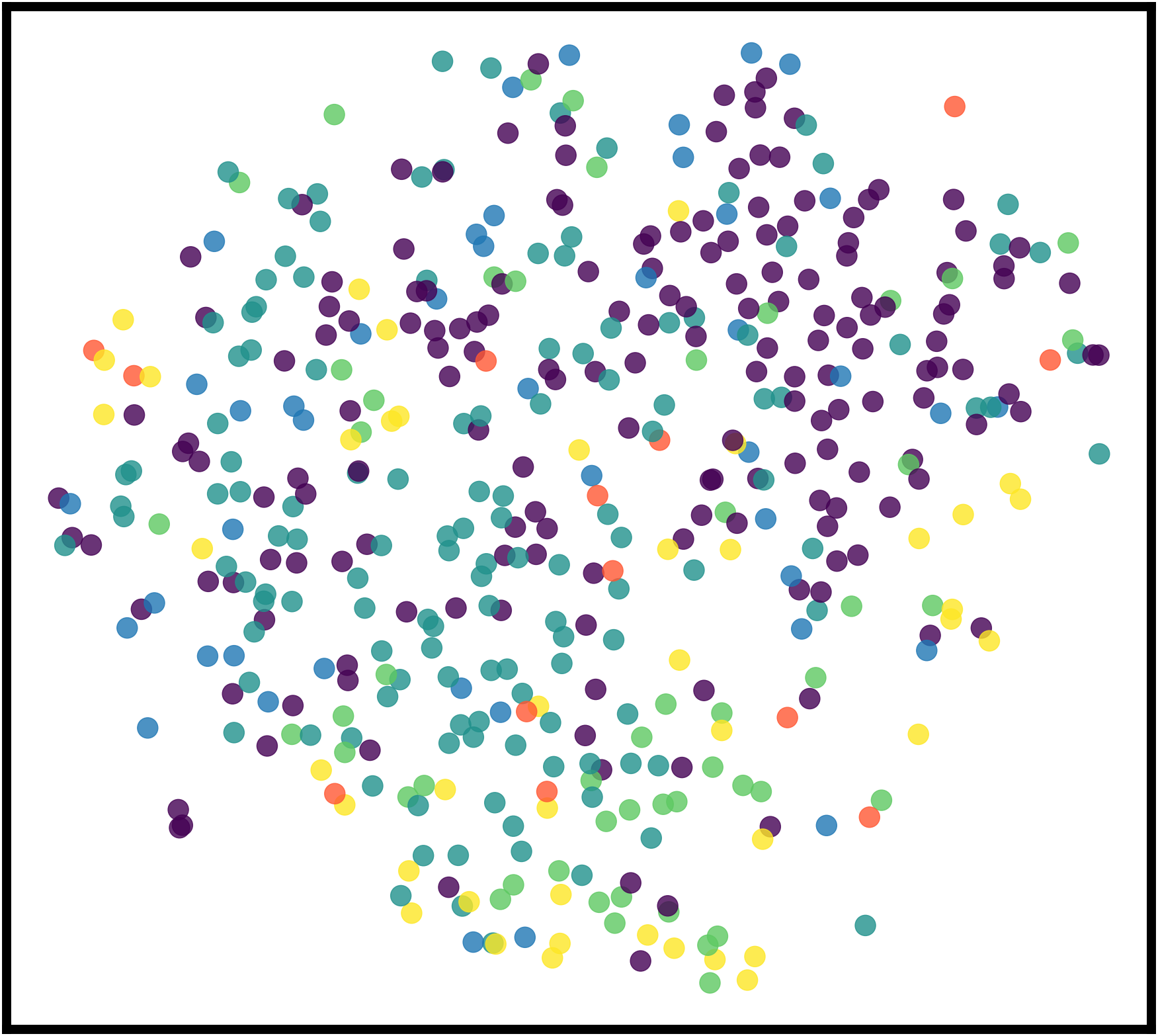} 
    } \hspace{-2.3ex}
\subfigure{\includegraphics[width=0.165\textwidth]{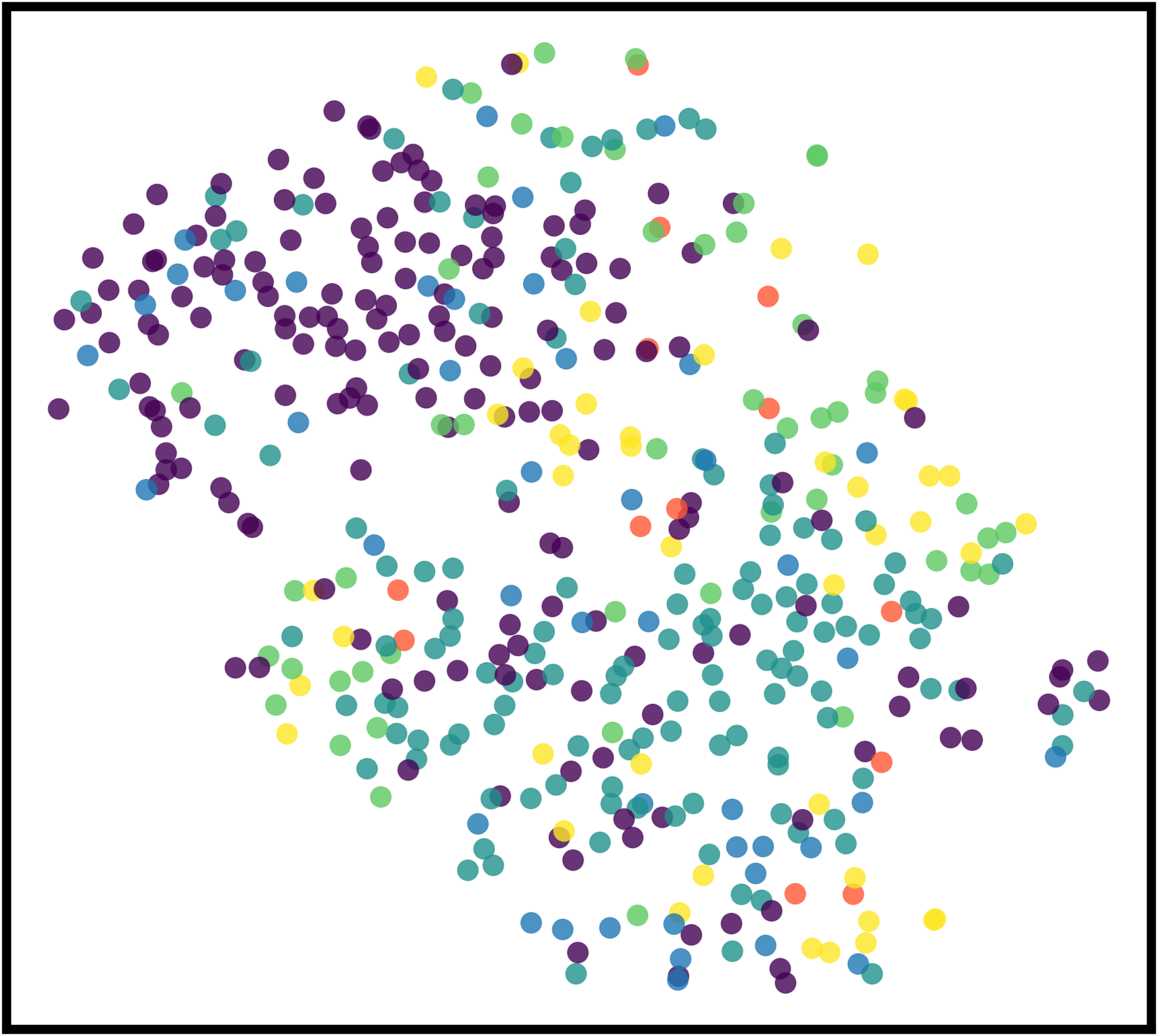}
    }\hspace{-1.5ex}  \subfigure{\includegraphics[width=0.165\textwidth]{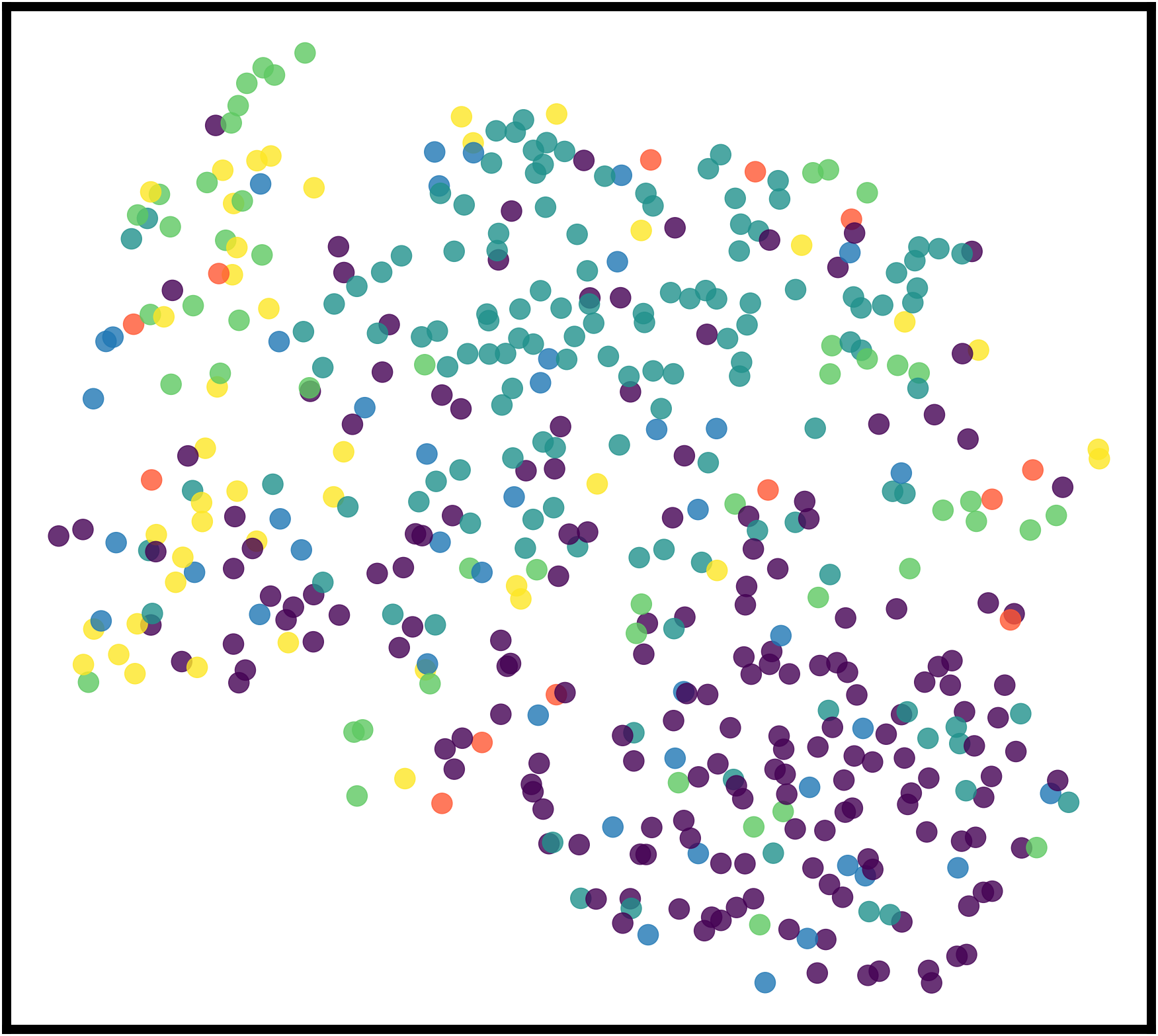}
    }\hspace{-1.5ex}
\subfigure{\includegraphics[width=0.165\textwidth]{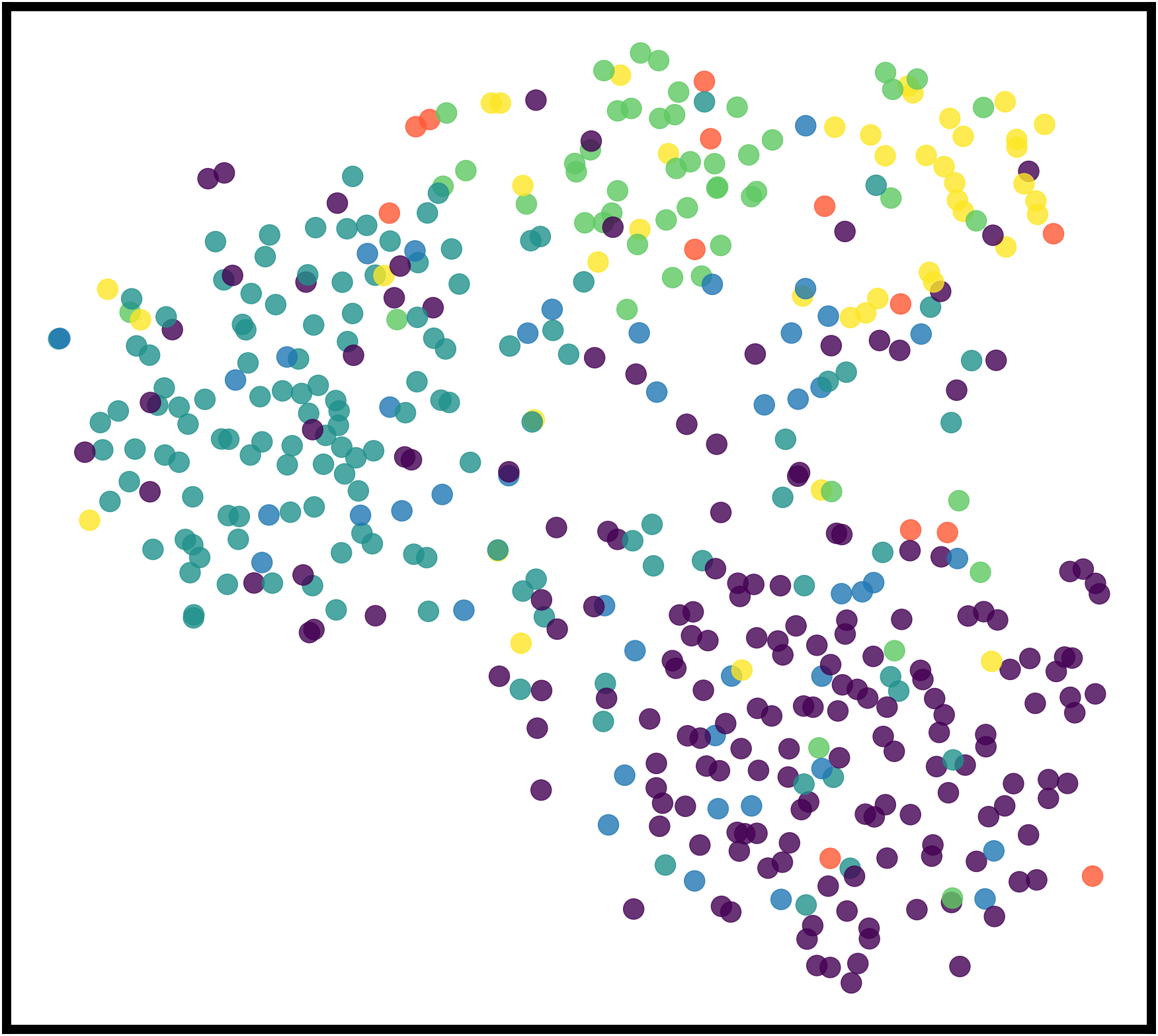}
    }\hspace{-1.5ex}  \subfigure{\includegraphics[width=0.165\textwidth]{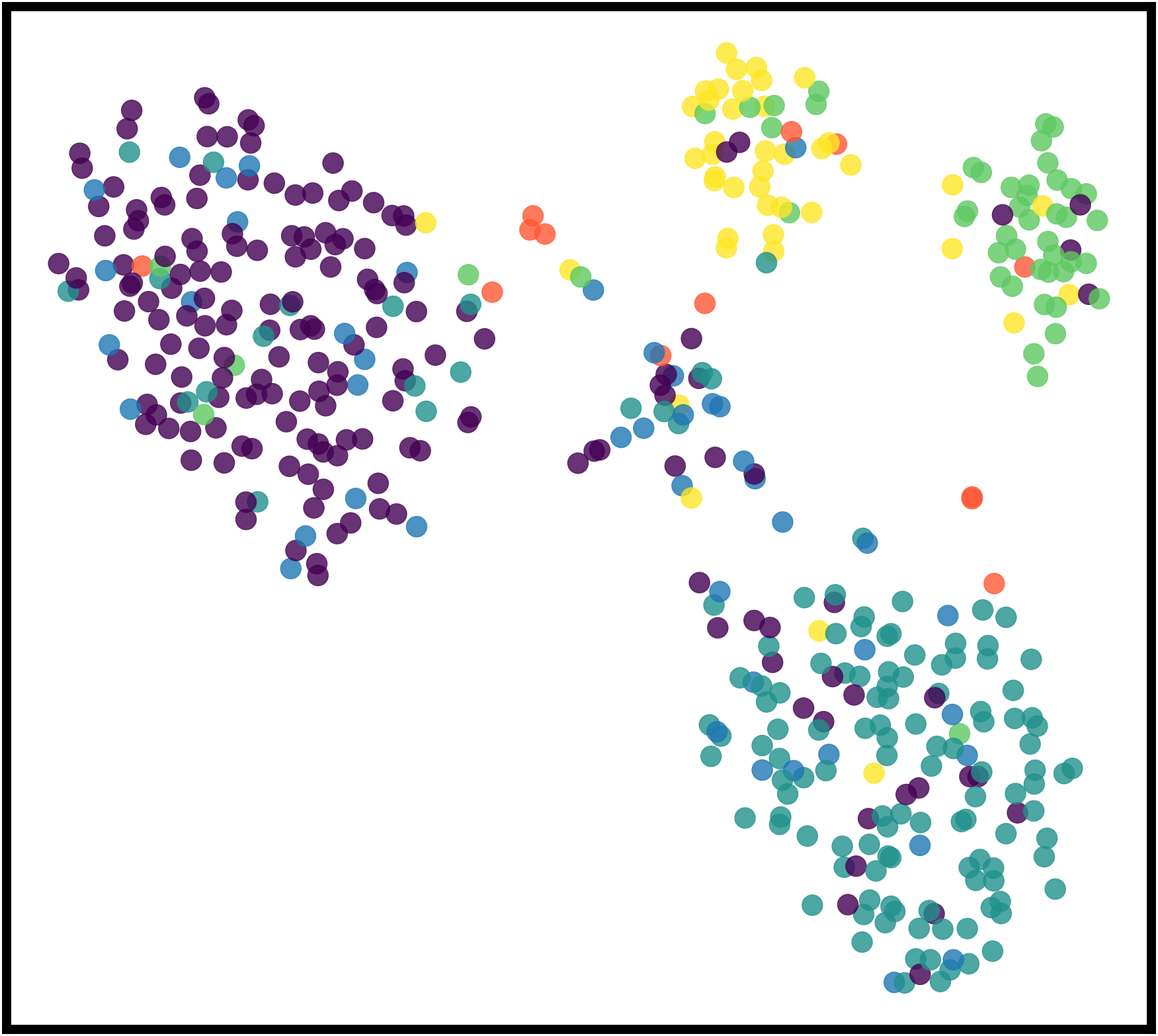}
    }\hspace{-1.5ex}
    \subfigure{\includegraphics[width=0.165\textwidth]{figures/pad/output_liuci0.png}
    }\hspace{-1.5ex}\\
\vspace{-0.2cm}
\setcounter{subfigure}{0} 
\hspace{-1.5ex}
\subfigure[t=250]{\includegraphics[width=0.165\textwidth]{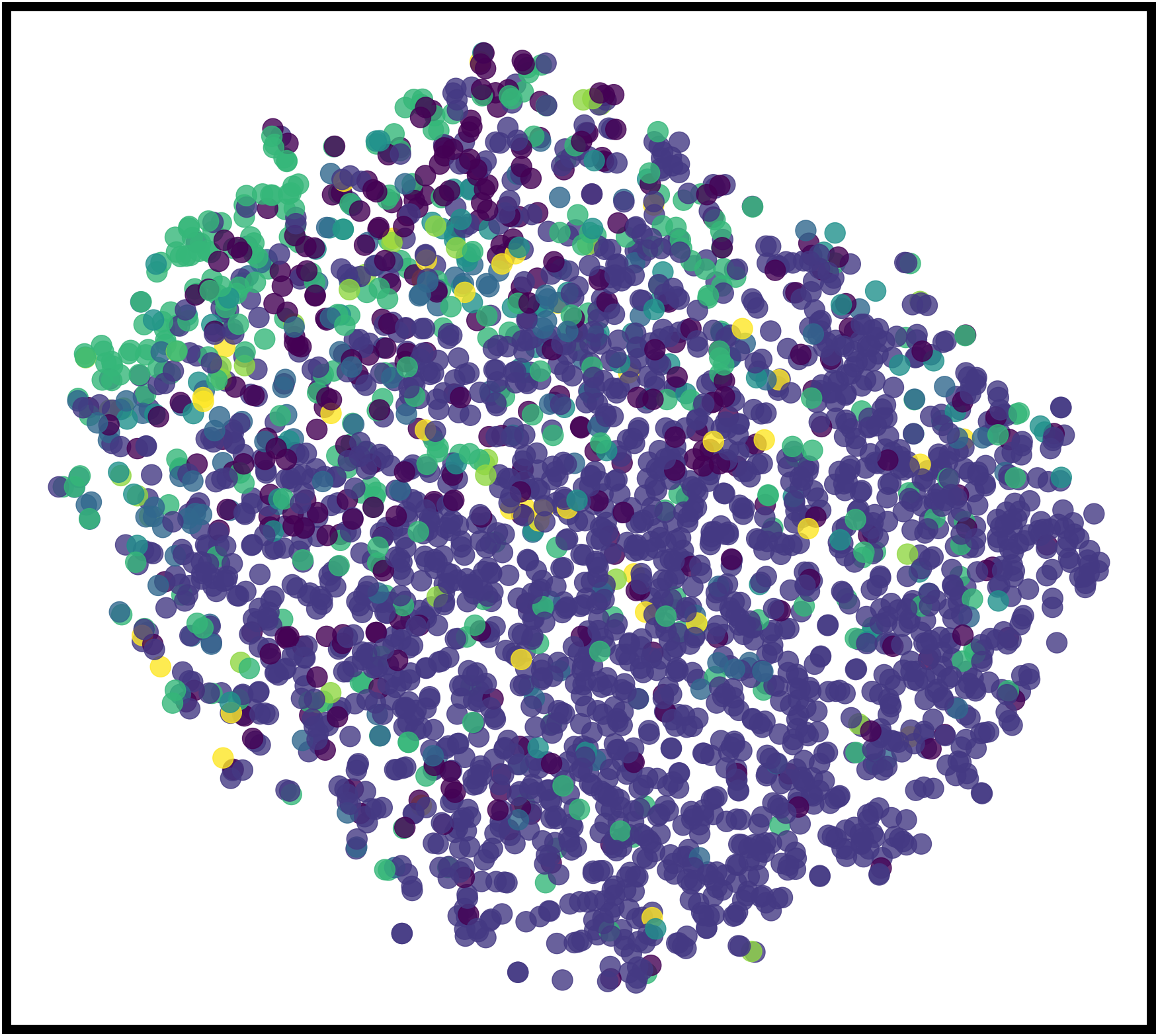}
    }\hspace{-1.5ex}
\subfigure[t=200]{\includegraphics[width=0.165\textwidth]{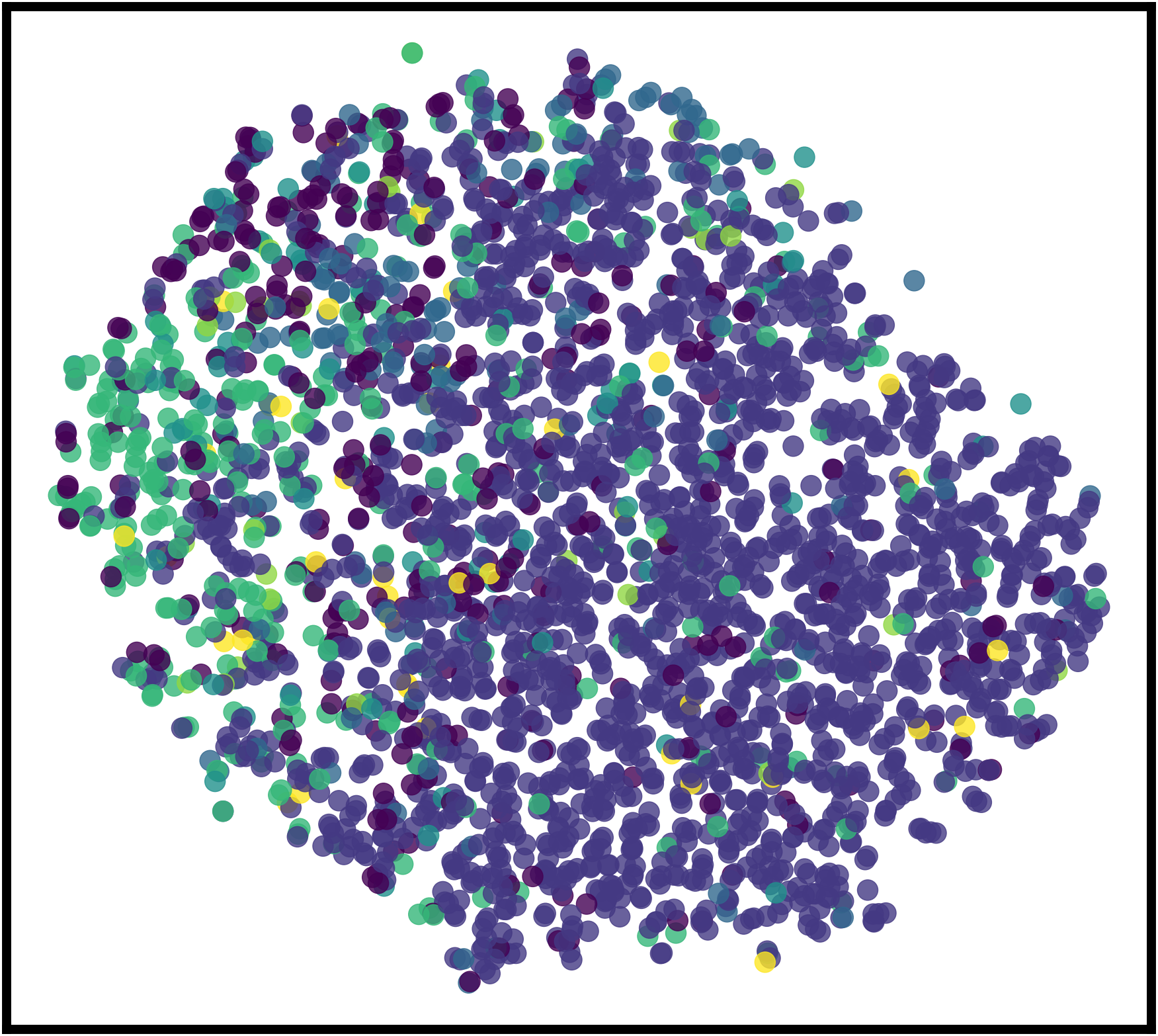}
    }\hspace{-1.5ex}
\subfigure[t=150]{\includegraphics[width=0.165\textwidth]{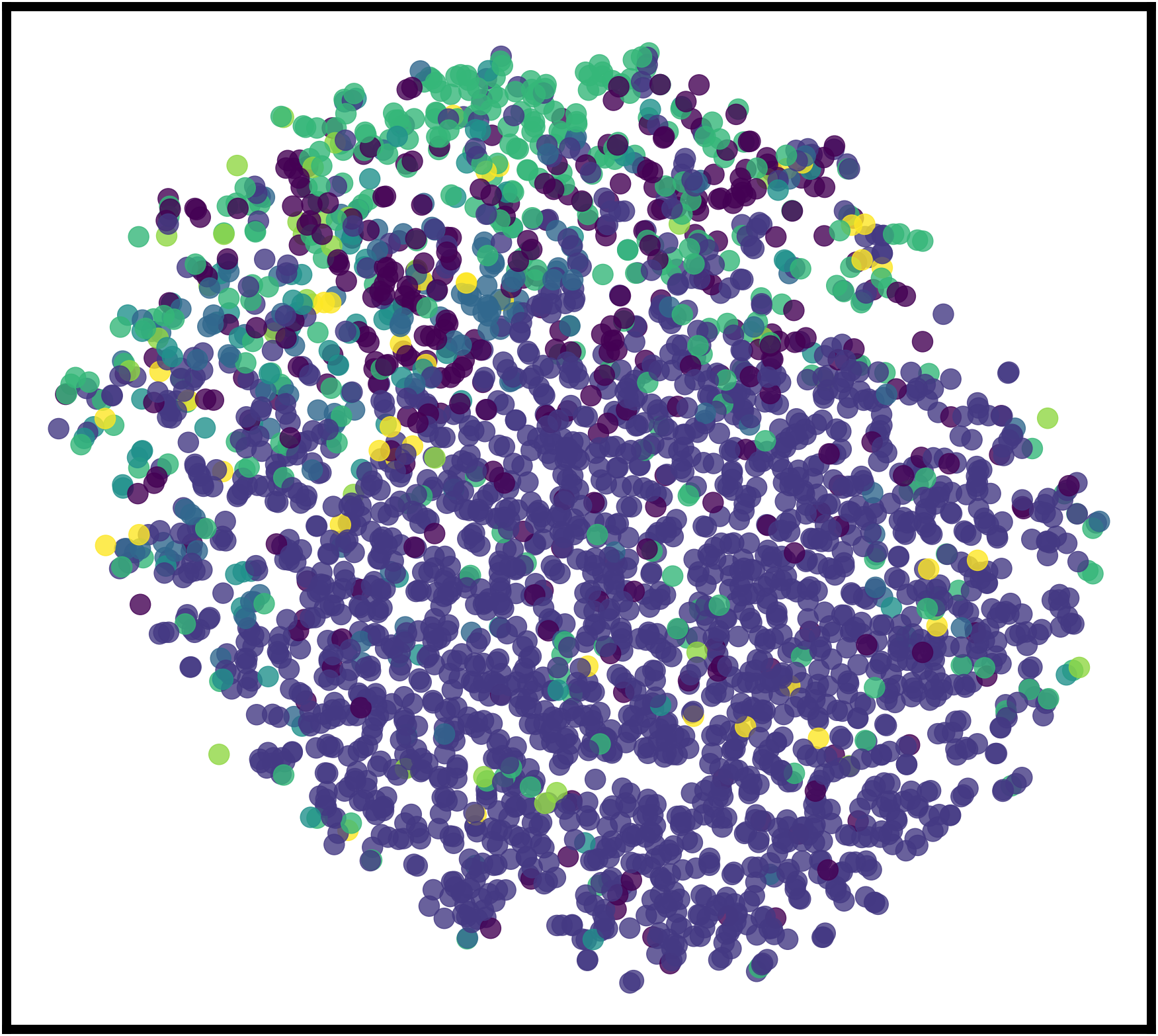}
    }\hspace{-1.5ex}
\subfigure[t=100]{\includegraphics[width=0.165\textwidth]{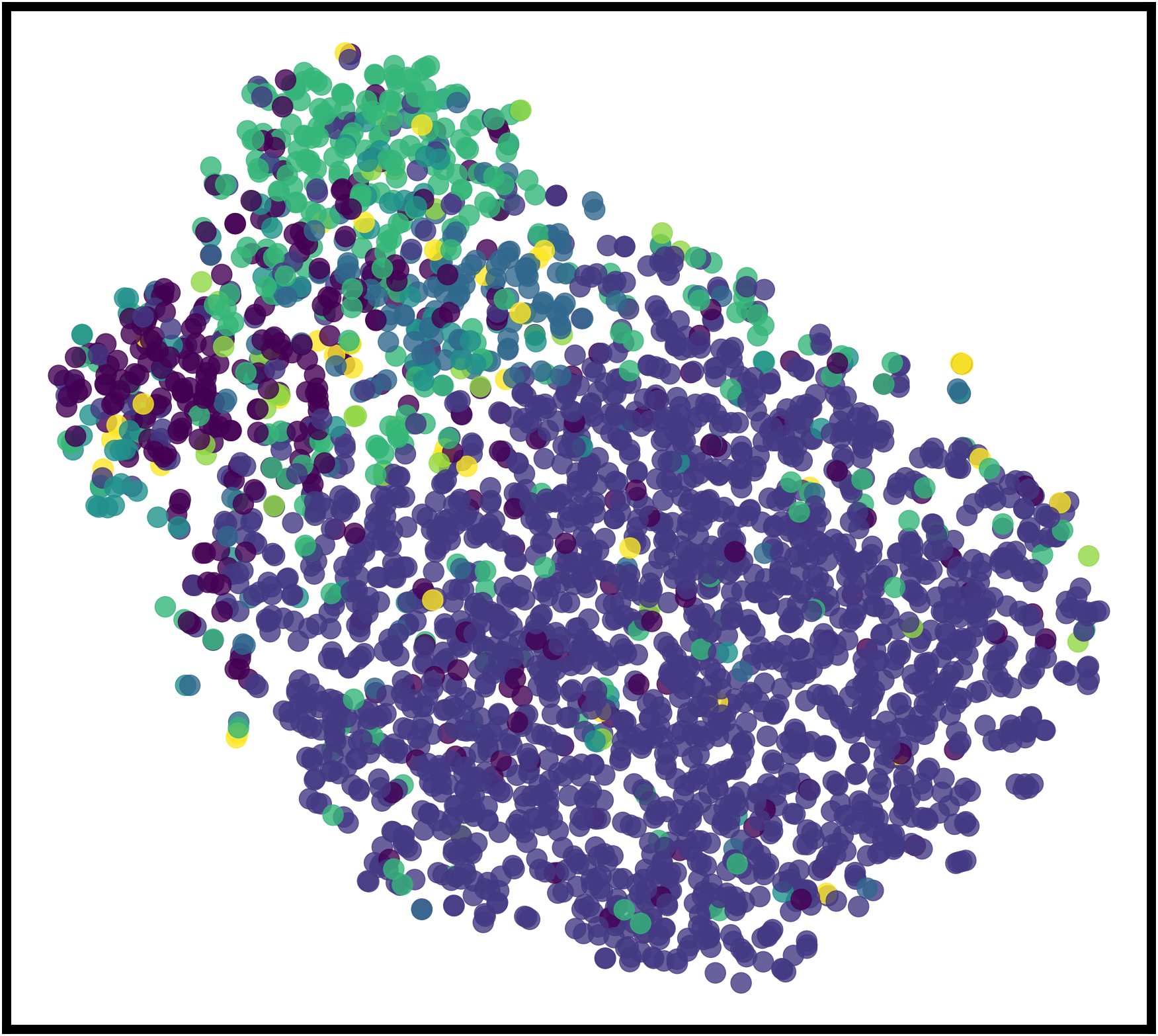}
    }\hspace{-1.5ex}
\subfigure[t=50]{\includegraphics[width=0.165\textwidth]{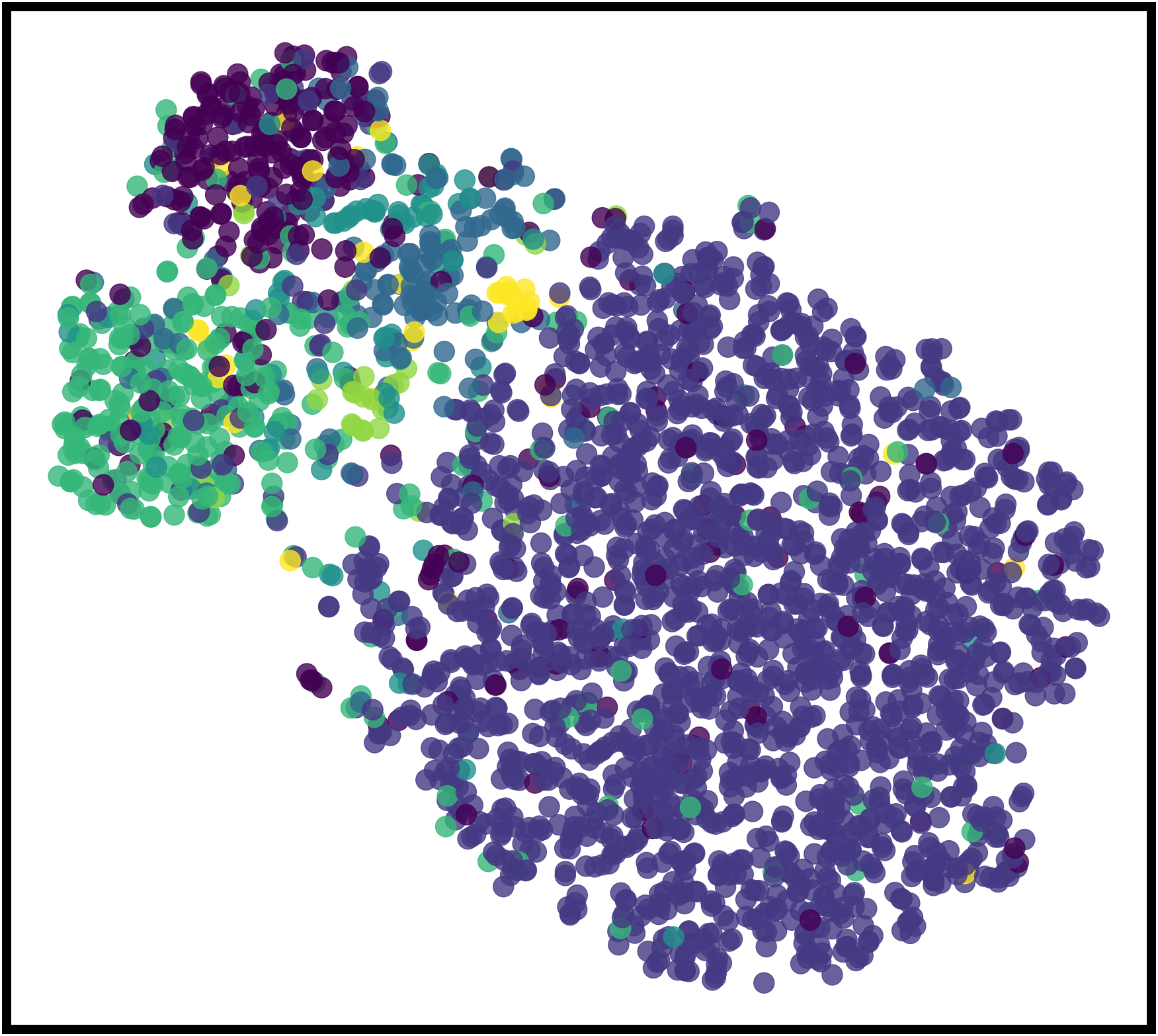}
    }\hspace{-1.4ex}
\subfigure[t=0]{\includegraphics[width=0.165\textwidth]{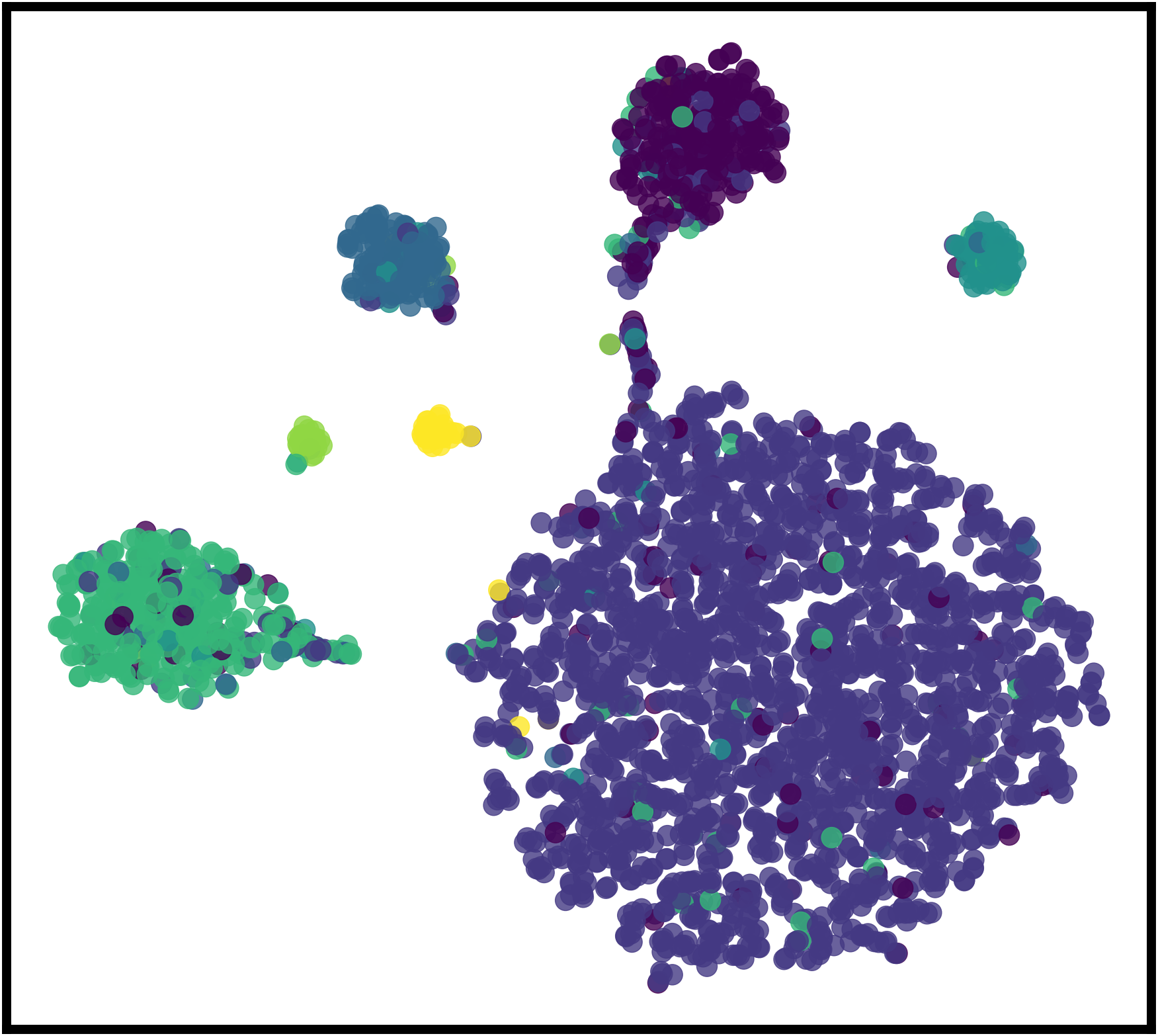}
    }\hspace{-1.5ex}
\caption{The t-SNE obtained from the denoised feature embedding by the diffusion reverse
process during inference on PAD-UFES and HAM10000 dataset, where $t$ is the current diffusion time step for
inference. As the time step encoding progresses, the noise is gradually removed, thereby
obtaining a clear distribution of classes.\label{convergence}}
\end{figure*}

To illustrate the numerical convergence of our ADPM during the reverse processing, we also used the t-SNE tool to visualize the denoised feature embeding at consecutive time steps. Both PAD-UFES and HAM10000 are used as shown in Figure \ref{convergence}, we can observe that as the denoising process progresses with time step encoding, our ADPM can gradually remove noise from the feature representations, thereby rendering the class distributions more distinct from the Gaussian distribution.

\subsection{Results and analysis}\label{results and analysis}


\textbf{PAD-UFES dataset.} We compared our ADPM against the state-of-the-art methods on the PAD-UFES dataset. Quantitative results in Table \ref{PADresult} show that our ADPM achieves the highest accuracy of 0.778 and precision of 0.752, indicating its effectiveness in classification tasks. The F1-score provides a balanced measure by harmonizing precision and recall, that is especially significant when dealing with classes of different sizes. As seen, PAD \cite{pacheco2020impact} gives the best F1-score, but its accuracy is much lower than ours. The reason is they utilized a subset of the dataset, which includes 1612 samples. The F1-score shows that our method has a significant improvement in classification accuracy compared to other methods. Therefore, it indicates that introducing anisotropic noise can effectively mitigate the test error caused by the insufficient number of samples in tail classes. 

\begin{table}[h]
\renewcommand{\arraystretch}{1.5}
\centering
\caption{Comparison results among state-of-the-art classification methods on PAD-UFES dataset.}\label{PADresult} 
\begin{tabular}{ccccc}
\toprule
Method & Accuracy  &F1-score & Precision & Recall\\
    \hline
    PAD\cite{pacheco2020impact} & 0.707$\pm$0.028 & 0.710$\pm$0.029 & 0.734$\pm$0.029 & 0.708$\pm$0.028\\
    ResNet-50\cite{he2016deep} & 0.671$\pm$0.041 & 0.678$\pm$0.037 & 0.720$\pm$0.041 & 0.670$\pm$0.041\\
    EfficientNet\cite{pmlr-v97-tan19a} & 0.745$\pm$0.022 & 0.651$\pm$0.044 & 0.695$\pm$0.045 & 0.640$\pm$0.039\\
    ViT-base\cite{dosovitskiy2020image} & 0.706$\pm$0.018 & 0.699$\pm$0.014 & 0.694$\pm$0.015 & \textbf{0.704$\pm$0.022}\\
    DiffMIC\cite{yang2023diffmic}  & 0.772$\pm$0.020 & 0.671$\pm$0.037 & 0.742$\pm$0.027 & 0.657$\pm$0.040 \\
    \hline
    \textbf{ADPM} & \textbf{0.778$\pm$0.011} & \textbf{0.701$\pm$0.019} & \textbf{0.752$\pm$0.034} & 0.682$\pm$0.024\\
\bottomrule
    \end{tabular}
\end{table}

Furthermore, we compared the ADPM with DiffMIC across each class to illustrate their predictive capabilities for minority classes. As shown in Table \ref{PADclassresult}, it can be observed that ADPM significantly improves the classifiation accuracy for Melanoma (MEL), Squamous Cell Carcinoma (SCC), Nevus (NEV), and Seborrheic Keratosis (SEK). Notably, our model improves the F1-score of Melanoma, which has the fewest samples, by more than 20\% compared to DiffMIC. The substantial improvements in the tail classes demonstrate the advantages of our approach in addressing the long-tail problem. Since anisotropic noise can balance the learning difficulty between head classes and tail classes, it leads to a slight decrease in the classification accuracy of head classes, such as Basal Cell Carcinoma (BCC). It explains why the difference of prediction accuracy between our ADPM and DiffMIC is not obvious. 

\begin{table}[h]
\caption{Comparison results between our ADPM and DiffMIC model on different classes of PAD-UFES dataset. 
\label{PADclassresult}}
\renewcommand{\arraystretch}{1.5}
\centering
\begin{tabular}{ccccc}
\toprule
\multirow{2}{*}{Category} & \multicolumn{2}{c}{DiffMIC \cite{yang2023diffmic}} & \multicolumn{2}{c}{ADPM} \\ 
\cmidrule(lr){2-3} \cmidrule(lr){4-5}
 & Accuracy  & F1-score & Accuracy  & F1-score \\ 
\hline 
MEL & 0.493$\pm$0.245 & 0.539$\pm$0.193 & \textbf{0.589$\pm$0.121} & \textbf{0.643$\pm$0.062} \\
SCC & 0.287$\pm$0.053 & 0.374$\pm$0.045 & \textbf{0.296$\pm$0.122} & \textbf{0.386$\pm$0.087} \\
BCC & \textbf{0.866$\pm$0.021} & \textbf{0.815$\pm$0.013} & 0.863$\pm$0.027 & 0.813$\pm$0.008 \\
ACK & 0.826$\pm$0.042 & 0.814$\pm$0.026 & \textbf{0.828$\pm$0.028} & \textbf{0.822$\pm$0.018} \\
NEV & 0.738$\pm$0.056 & 0.751$\pm$0.008 & \textbf{0.761$\pm$0.035} & \textbf{0.771$\pm$0.009} \\
SEK & 0.733$\pm$0.069 & 0.723$\pm$0.041 & \textbf{0.769$\pm$0.050} & \textbf{0.773$\pm$0.037} \\
\bottomrule
\end{tabular}
\end{table}

\textbf{HAM10000 dataset.} We compared the classification performance of ADPM with state-of-the-art medical image classification methods on the dermoscopic image dataset. As shown in Table \ref{HAMresult}, our proposed method achieves an accuracy of 0.906 and the highest F1-score of 0.846, which is significantly higher than all other comparison methods. The higher F1-score makes it more reliable for classification tasks involving imbalanced data.
\begin{table*}[h]
\centering
\renewcommand{\arraystretch}{1.5}
\setlength{\tabcolsep}{0.4pt}
\caption{Comparison results among state-of-the-art classification methods on HAM10000 dataset.}\label{HAMresult}
    \begin{tabular}{c|ccccccc|c}
    \toprule
    Method & LDAM\cite{cao2019learning}  &OHEM\cite{shrivastava2016training} & MTL\cite{haofu2017deep} & DANIL\cite{gong2020distractor} & CL\cite{marrakchi2021fighting} & ProCo\cite{yang2022proco} & DiffMIC\cite{yang2023diffmic} & \textbf{ADPM}\\
    \hline
    Accuracy & 0.857 & 0.818 & 0.811 & 0.825 & 0.865 & 0.887 & \textbf{0.906} & \textbf{0.906}  \\
    F1-score & 0.734 & 0.660 & 0.667 & 0.674 & 0.739 & 0.763 & 0.816 & \textbf{0.846}\\
    \bottomrule
    \end{tabular}
\end{table*}

Table \ref{HAMclassresult} compared our ADPM against DiffMIC across various categories in the HAM10000 dataset, including accuracy and F1-score. We focus on the tailed classes such as Actinic Keratosis/Bowen's Disease (AKIEC), Dermatofibroma (DF), and Vascular Lesions (VASC). For these tailed classes, our model consistently achieves higher F1-score. For example, for the DF class, ADPM significantly improves the F1-score while maintaining accuracy, showcasing a better balance between precision and recall. The improvement for the DF class is the most notable, with accuracy increasing from 0.688 to 0.844 and the F1-score increasing from 0.733 to 0.853. 
These results emphasize the robust performance of our model in classification tasks involving imbalanced datasets, ensuring reliable and accurate predictions across all classes, especially those with fewer samples.

\begin{table}[h]
\renewcommand{\arraystretch}{1.5}
\setlength{\tabcolsep}{10pt}
\caption{Comparison to DiffMIC on each class of HAM10000 dataset\label{HAMclassresult}}
\centering
\begin{tabular}{cccccc}
\toprule
\multirow{2}{*}{Category} & \multicolumn{2}{c}{DiffMIC \cite{yang2023diffmic}} & \multicolumn{1}{l}{} & \multicolumn{2}{c}{ADPM} \\
\cmidrule(r){2-3} \cmidrule(r){5-6}
 & Accuracy  & F1-score  & \multicolumn{1}{l}{} & Accuracy  & F1-score  \\ 
\hline 
MEL & 0.669 & 0.733 & & \textbf{0.693} & \textbf{0.748} \\
NV & \textbf{0.971} & \textbf{0.956} & & 0.968 & 0.953 \\
BCC & 0.836 & 0.838 & & \textbf{0.842} & \textbf{0.869} \\
AKIEC & \textbf{0.716} & 0.783 & & \textbf{0.716} & \textbf{0.787 }\\
BKL & 0.805 & 0.805 & & \textbf{0.833} & \textbf{0.808} \\
DF & 0.688 & 0.733 & & \textbf{0.844} & \textbf{0.853} \\
VASC & \textbf{0.872} & 0.861 & & 0.821 & \textbf{0.901} \\
\bottomrule
\end{tabular}
\end{table}

\textbf{SCIN dataset.}  
Since patients in the SCIN dataset may have multiple dermatological diagnoses, we selected the diagnosis with the highest weight as the label for each case. Additionally, due to the wide variety of dermatological conditions diagnosed by different clinicians, and in line with previous dermatological research by Google Health \cite{liu2020deep}, we selected cases of the 26 most prevalent skin diseases in clinical practice as the dataset for our model.
We evaluation the performance of our ADPM by comparing it with MobileNet \cite{Sandler_2018_CVPR}, InceptionNet \cite{szegedy2017inception}, ViT-based model \cite{dosovitskiy2020image} and DiffMIC \cite{yang2023diffmic}. 
We used the five-fold cross validation for the comparison methods, the results of which are displayed in Table \ref{SCINresult}. Compared to InceptionNet used by Google Health team \cite{liu2020deep}, our ADPM gives the best performances in both classification accuracy and F1-score with more than 1\% higher values than DiffMIC method. The results indicate that as the number of categories increases, image classification becomes more challenging. The F1-score reflects the difficulty in classifying tail classes, leading to lower overall accuracy in the dataset.

\begin{table}[h] 
\renewcommand{\arraystretch}{1.5}
\centering
\caption{Comparison results among state-of-the-art classification methods on SCIN dataset.}\label{SCINresult}
    \begin{tabular}{ccccc}
    \toprule
    Method & Accuracy  & F1-score & Precision  & Recall \\
    \hline
    MobileNet \cite{Sandler_2018_CVPR} & ~~0.583$\pm$0.018 ~~&~~ 0.384$\pm$0.053~~ & ~~0.422$\pm$0.086~~ & ~~0.363$\pm$0.048~~ \\
    InceptionNet \cite{szegedy2017inception} & 0.615$\pm$0.019 & 0.411$\pm$0.046 & 0.446$\pm$0.060 & 0.397$\pm$0.035 \\
    ViT-base \cite{dosovitskiy2020image} & 0.625$\pm$0.027 & 0.482$\pm$0.054 & 0.571$\pm$0.058 & 0.447$\pm$0.052 \\
    DiffMIC \cite{yang2023diffmic} & 0.634$\pm$0.015 & 0.491$\pm$0.019 & 0.599$\pm$0.038 & 0.452$\pm$0.021 \\
    \hline
    \textbf{ADPM} & \textbf{0.645$\pm$0.012} & \textbf{0.504$\pm$0.033} & \textbf{0.606$\pm$0.037} & \textbf{0.462$\pm$0.025} \\ 
    \bottomrule
    \end{tabular}
\end{table}

\begin{figure*}[h]
    \centering
    \includegraphics[scale=0.35]{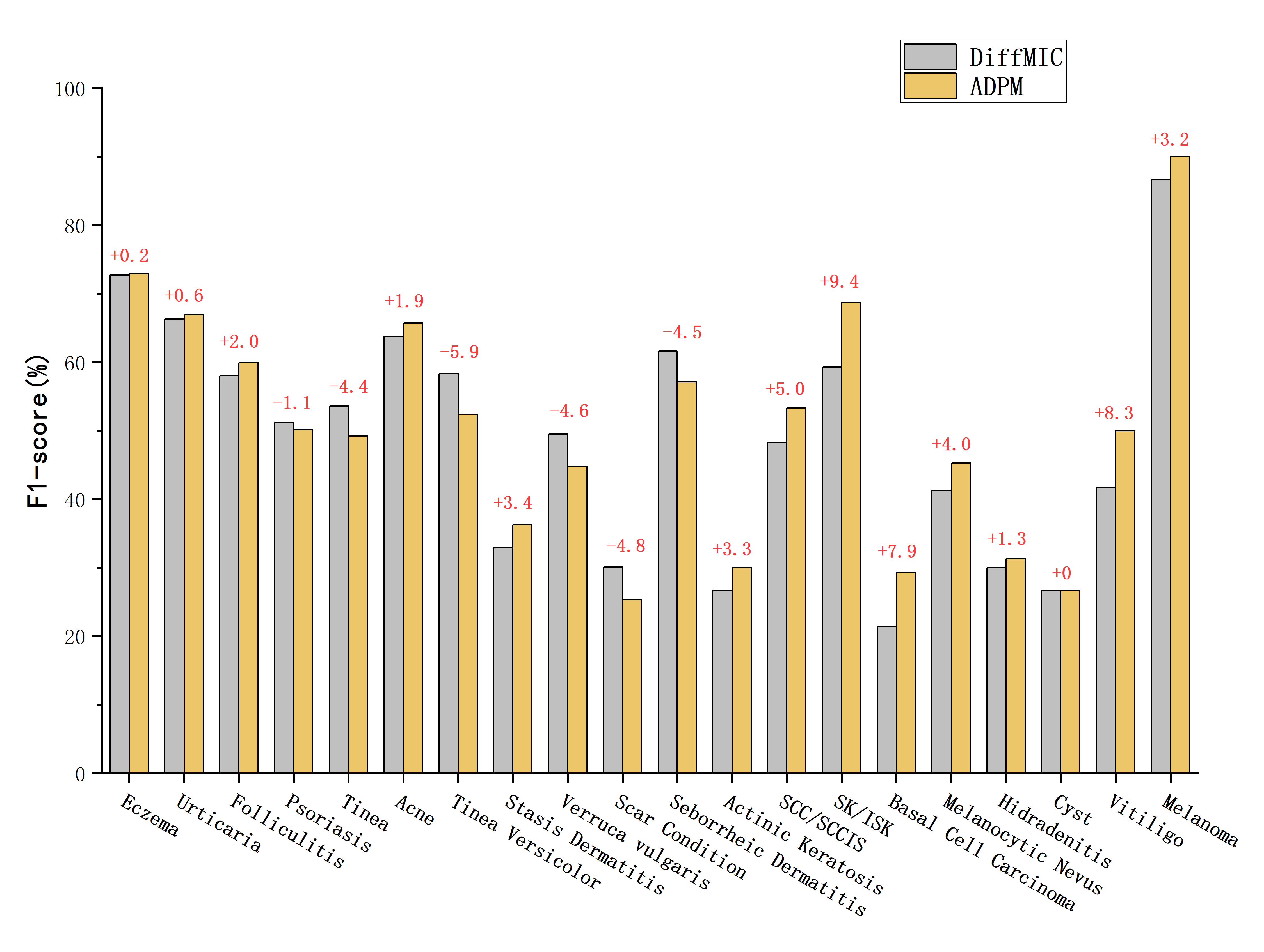}
    \vspace{-0.6cm}
    \caption{Comparison results between our ADPM and DiffMIC model across different categories of SCIN dataset, where the categories are sorted from left to right based on the number of samples.}
    \label{scinclass}
\end{figure*}

As shown in Figure \ref{scinclass}, our model shows a significant improvement in classification accuracy across various tail classes, where the categories are sorted from left to right based on the number of samples. Tail categories, such as Actinic Keratosis, SCC/SCCIS, and Basal Cell Carcinoma, with sample sizes of 21, 16, and 13, respectively, show significantly higher classification accuracy compared to DiffMIC, which is of great importance for the diagnosis of rare but critical diseases. On the other hand, head region classes, such as Eczema, Urticaria, Acne, and Stasis Dermatitis, are relatively less sensitive to noise. Even with the addition of a small amount of noise, our model still achieves better classification accuracy than DiffMIC. However, for the middle classes between the head and tail categories, classification accuracy declines compared to DiffMIC due to noise variations. The increasing number of categories challenges classification methods and the estimation of anisotropic noise in diffusion models. 
We believe that grouping the categories will help estimate more appropriate noise intensities, further improving our model's classification accuracy. It is a key area of exploration in our future research.

\begin{figure*}[h]
    \centering
    \includegraphics[scale=0.57]{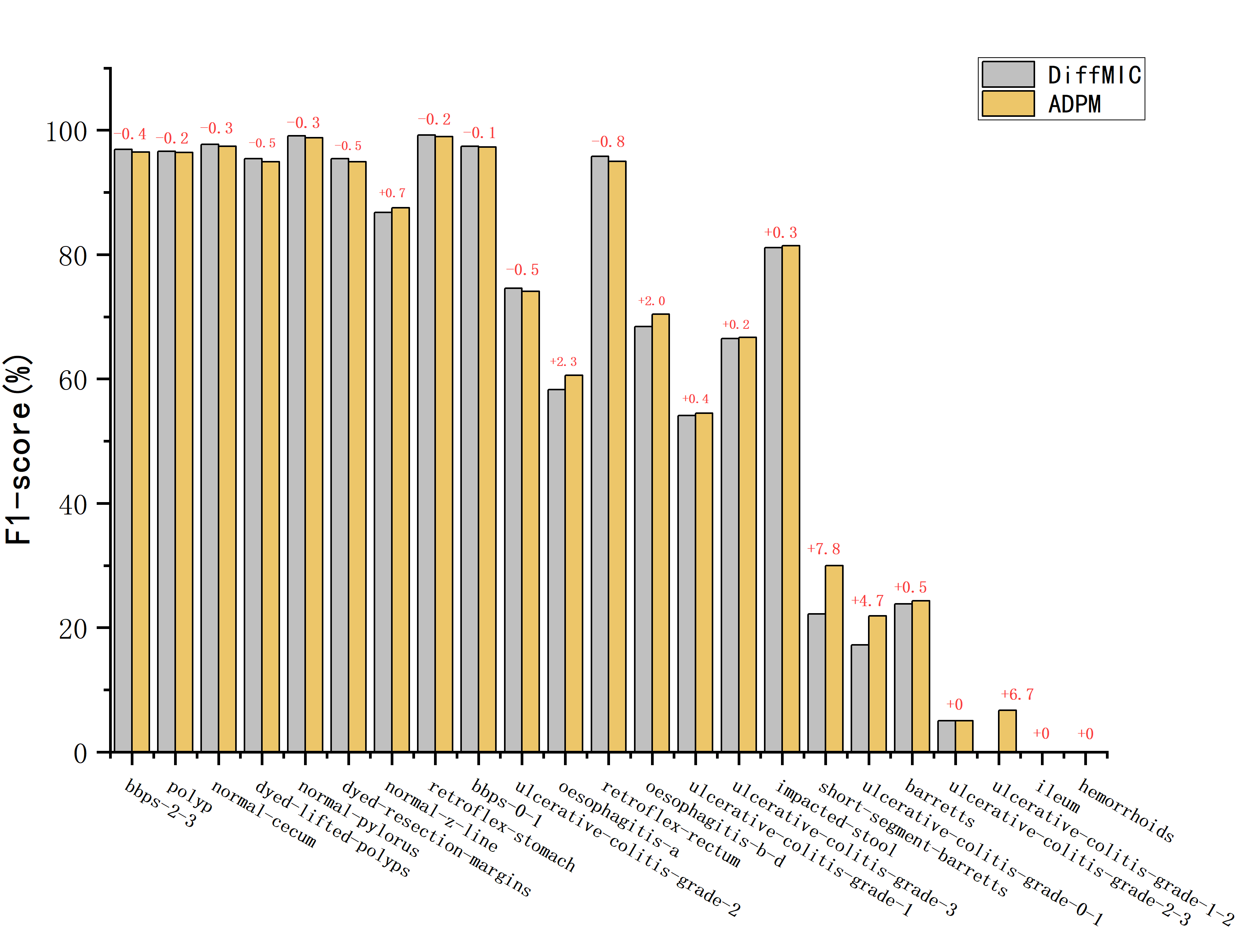}
    \vspace{-0.6cm}
    \caption{Comparison results between our ADPM and DiffMIC model across different categories of Hyper-Kvasir dataset, where the categories are sorted from left to right based on the number of samples.}
    \label{hyperclass}
\end{figure*}

\textbf{Hyper-Kvasir dataset.}
In addition to skin lesion classfication, we conducted comparative experiments on the gastrointestinal Hyper-Kvasir dataset to validate the effectiveness of our method. We not only employed classification models based on ResNet, MobileNet, DenseNet, and InceptionNet, but also compared them with one of the latest class-imbalanced classification models, PVT \cite{wang2022pvt}. Note that for the MobileNet, InceptionNet and PVT, we used the dynamically weighted balanced (DWB) loss \cite{9324926}, which incorporates a modulation factor that allows the network to focus more on samples with lower prediction values and those from minority classes. Considering that the model cannot differentiate some tail classes in Hyper data due to insufficient samples, we remove the influence of these samples when adding anisotropic noise. As shown in Table \ref{Hyperresult}, compared to other methods, our model achieves the highest F1-score and recall while maintaining high accuracy and precision. It demonstrates the superior ability to handle class-imbalanced problems. 

\begin{table}[t]
\renewcommand{\arraystretch}{1.5}
\centering
\caption{Comparison results among state-of-the-art classification methods on Hyper-Kvasir dataset.}\label{Hyperresult}
    \begin{tabular}{ccccc}
    \toprule
    Method & Accuracy  & F1-score & Precision  & Recall \\
    \hline
    ResNet-50 \cite{he2016deep} & ~~0.853$\pm$0.005~~ & ~~0.546$\pm$0.007~~ & ~~0.547$\pm$0.007~~ & ~~0.552$\pm$0.009~~\\
    MobileNet \cite{Sandler_2018_CVPR} & 0.883$\pm$0.005 & 0.586$\pm$0.018 & 0.592$\pm$0.031 & 0.583$\pm$0.015\\
    DenseNet-161 \cite{huang2017densely} & 0.907$\pm$0.006 & 0.619$\pm$0.024 & 0.635$\pm$0.036 & 0.618$\pm$0.028\\
    InceptionNet \cite{szegedy2017inception} & \textbf{0.908$\pm$0.002} & 0.612$\pm$0.014 & 0.630$\pm$0.033 & 0.612$\pm$0.017\\
    PVT-v2-B1 \cite{wang2022pvt} & 0.900$\pm$0.007 & 0.614$\pm$0.031 & 0.627$\pm$0.042 & 0.616$\pm$0.025 \\
    DiffMIC \cite{yang2023diffmic} & 0.901$\pm$0.006 & 0.624$\pm$0.020 & 0.645$\pm$0.043 & 0.621$\pm$0.016 \\
    \hline
    \textbf{ADPM} & 0.904$\pm$0.003 & \textbf{0.631$\pm$0.019} & \textbf{0.655$\pm$0.031} & \textbf{0.626$\pm$0.022} \\ 
    \bottomrule
    \end{tabular}
\end{table}

Besides, we compared the classification performance across all categories with the DiffMIC method, as shown in Figure \ref{hyperclass}. For head classes, both methods achieved satisfactory classification accuracy, while our proposed method significantly improved the classification accuracy for tail classes. Thus, by controlling the diffusion speed of different categories, it effectively balances the influence of underrepresented classes, thereby enhancing its generalization ability in complex classification scenarios.  However, for the two categories with very few samples, ileum and hemorrhoids, with only 9 and 6 samples respectively, neither method could classify them, effectively. Thus, we need to incorporate other techniques, such as resampling, to address this extremely challenging task.

\section{Conclusion and Discussion}\label{Conclusion}
In this work, we have presented a theoretically sound diffusion probabilitic model by introducing anisotropic distributed noise to address the problem of long-tailed  data classification. Relying on the nonlinear relationship between the generalization error and the sample size, we  balanced the distribution of long-tailed classes by introducing anisotropic noise during the forward diffusion process, thereby achieving effective control the diffusion speed of the long-tailed classes. We theoretically proposed an effective method for controlling the generalization error in long-tailed data classification problems, which can be used to design resampling strategies to balance the samples in long-tailed distributed data. Numerical experiments on typical imbalanced image classification problems were conducted to demonstrate the effectiveness of the proposed method.  

The proposed anisotropic diffusion probabilistic model is built on the assumption that the training dataset and test dataset share similar distributions. In practice, whether the data distribution of training and testing samples is consistent depends on the specific application scenario and data collection methods. We leave the issue of choosing the diffusion speed in the probabilistic model when the class distribution in the test scenario is unknown for future study. Additionally, we determined the choice of noise parameters by comparing classification accuracy under different parameter combinations on the dataset. This is a feasible method, but the theoretically optimal values of noise parameters still require further investigation. Additionally, for classification problems with many categories, clustering categories based on inter-class distances and then re-estimating anisotropic noise is a promising direction for exploration.

\section*{Acknowledgments}
The work is supported by the National Natural Science Foundation of China (NSFC 12071345).

\bibliographystyle{siamplain}
\bibliography{references}

\begin{thebibliography}{10}

\bibitem{bartlett2002rademacher}
{\sc P.~L. Bartlett and S.~Mendelson}, {\em Rademacher and gaussian complexities: Risk bounds and structural results}, Journal of Machine Learning Research, 3 (2002), pp.~463--482.

\bibitem{Borgli2020}
{\sc H.~Borgli, V.~Thambawita, P.~H. Smedsrud, and Hicks}, {\em Hyperkvasir, a comprehensive multi-class image and video dataset for gastrointestinal endoscopy}, Scientific Data, 7 (2020), p.~283.

\bibitem{doi:10.1137/22M1503579}
{\sc N.~D. Bruhin and B.~Davies}, {\em Bioinspired random projections for robust, sparse classification}, SIAM Journal on Imaging Sciences, 15 (2022), pp.~1833--1850.

\bibitem{cao2019learning}
{\sc K.~Cao, C.~Wei, A.~Gaidon, N.~Arechiga, and T.~Ma}, {\em Learning imbalanced datasets with label-distribution-aware margin loss}, in Advances in Neural Information Processing Systems, vol.~32, 2019.

\bibitem{chawla2002smote}
{\sc N.~V. Chawla, K.~W. Bowyer, L.~O. Hall, and W.~P. Kegelmeyer}, {\em Smote: synthetic minority over-sampling technique}, Journal of Artificial Intelligence Research, 16 (2002), pp.~321--357.

\bibitem{10.1007/978-3-030-65414-6_9}
{\sc H.-P. Chou, S.-C. Chang, J.-Y. Pan, W.~Wei, and D.-C. Juan}, {\em Remix: Rebalanced mixup}, in Computer Vision -- ECCV 2020 Workshops, A.~Bartoli and A.~Fusiello, eds., Cham, 2020, Springer International Publishing, pp.~95--110.

\bibitem{dosovitskiy2020image}
{\sc A.~Dosovitskiy, L.~Beyer, A.~Kolesnikov, D.~Weissenborn, X.~Zhai, T.~Unterthiner, M.~Dehghani, M.~Minderer, G.~Heigold, S.~Gelly, J.~Uszkoreit, and N.~Houlsby}, {\em An image is worth 16$x$16 words: Transformers for image recognition at scale}, in International Conference on Learning Representations, 2021, \url{https://openreview.net/forum?id=YicbFdNTTy}.

\bibitem{9324926}
{\sc K.~R.~M. Fernando and C.~P. Tsokos}, {\em Dynamically weighted balanced loss: Class imbalanced learning and confidence calibration of deep neural networks}, IEEE Transactions on Neural Networks and Learning Systems, 33 (2022), pp.~2940--2951.

\bibitem{frisch2023synthesising}
{\sc Y.~Frisch, M.~Fuchs, A.~Sanner, F.~A. Ucar, M.~Frenzel, J.~Wasielica-Poslednik, A.~Gericke, F.~M. Wagner, T.~Dratsch, and A.~Mukhopadhyay}, {\em Synthesising rare cataract surgery samples with guided diffusion models}, in International Conference on Medical Image Computing and Computer-Assisted Intervention, Springer, 2023, pp.~354--364.

\bibitem{galdran2021balanced}
{\sc A.~Galdran, G.~Carneiro, and M.~A. Gonz{\'a}lez~Ballester}, {\em Balanced-mixup for highly imbalanced medical image classification}, in Medical Image Computing and Computer Assisted Intervention -- MICCAI 2021, Springer, 2021, pp.~323--333.

\bibitem{gong2020distractor}
{\sc L.~Gong, K.~Ma, and Y.~Zheng}, {\em Distractor-aware neuron intrinsic learning for generic 2d medical image classifications}, in Medical Image Computing and Computer Assisted Intervention -- MICCAI 2020, Springer, 2020, pp.~591--601.

\bibitem{han2005borderline}
{\sc H.~Han, W.-Y. Wang, and B.-H. Mao}, {\em Borderline-smote: A new over-sampling method in imbalanced data sets learning}, in Advances in Intelligent Computing, Berlin, Heidelberg, 2005, Springer Berlin Heidelberg, pp.~878--887.

\bibitem{han2024latent}
{\sc P.~Han, C.~Ye, J.~Zhou, J.~Zhang, J.~Hong, and X.~Li}, {\em Latent-based diffusion model for long-tailed recognition}, in Proceedings of the IEEE/CVF Conference on Computer Vision and Pattern Recognition, 2024, pp.~2639--2648.

\bibitem{han2022card}
{\sc X.~Han, H.~Zheng, and M.~Zhou}, {\em Card: Classification and regression diffusion models}, Advances in Neural Information Processing Systems, 35 (2022), pp.~18100--18115.

\bibitem{haofu2017deep}
{\sc L.~Haofu and J.~Luo}, {\em A deep multi-task learning approach to skin lesion classification}, in Workshops at the Thirty-First AAAI Conference on Artificial Intelligence, 2017.

\bibitem{he2009learning}
{\sc H.~He and E.~A. Garcia}, {\em Learning from imbalanced data}, IEEE Transactions on Knowledge and Data Engineering, 21 (2009), pp.~1263--1284.

\bibitem{he2016deep}
{\sc K.~He, X.~Zhang, S.~Ren, and J.~Sun}, {\em {Deep Residual Learning for Image Recognition}}, in Proceedings of the IEEE Conference on Computer Vision and Pattern Recognition, 2016, pp.~770--778.

\bibitem{hestness2017deep}
{\sc J.~Hestness, S.~Narang, N.~Ardalani, G.~Diamos, H.~Jun, H.~Kianinejad, M.~M.~A. Patwary, Y.~Yang, and Y.~Zhou}, {\em Deep learning scaling is predictable, empirically}, arXiv preprint arXiv:1712.00409,  (2017).

\bibitem{ho2020denoising}
{\sc J.~Ho, A.~Jain, and P.~Abbeel}, {\em {Denoising Diffusion Probabilistic Models}}, Advances in Neural Information Processing Systems, 33 (2020), pp.~6840--6851.

\bibitem{ho2022cascaded}
{\sc J.~Ho, C.~Saharia, W.~Chan, D.~J. Fleet, M.~Norouzi, and T.~Salimans}, {\em Cascaded diffusion models for high fidelity image generation}, Journal of Machine Learning Research, 23 (2022), pp.~1--33.

\bibitem{huang2017densely}
{\sc G.~Huang, Z.~Liu, L.~Van Der~Maaten, and K.~Q. Weinberger}, {\em {Densely Connected Convolutional Networks}}, in Proceedings of the IEEE Conference on Computer Vision and Pattern Recognition, 2017, pp.~4700--4708.

\bibitem{jain2016structural}
{\sc A.~Jain, A.~R. Zamir, S.~Savarese, and A.~Saxena}, {\em {Structural-RNN: Deep Learning on Spatio-Temporal Graphs}}, in Proceedings of the IEEE Conference on Computer Vision and Pattern Recognition, 2016, pp.~5308--5317.

\bibitem{jamal2020rethinking}
{\sc M.~A. Jamal, M.~Brown, M.-H. Yang, L.~Wang, and B.~Gong}, {\em {Rethinking Class-Balanced Methods for Long-Tailed Visual Recognition from a Domain Adaptation Perspective}}, in Proceedings of the IEEE/CVF Conference on Computer Vision and Pattern Recognition, 2020, pp.~7610--7619.

\bibitem{jeatrakul2010classification}
{\sc P.~Jeatrakul, K.~W. Wong, and C.~C. Fung}, {\em {Classification of Imbalanced Data by Combining the Complementary Neural Network and SMOTE Algorithm}}, in Neural Information Processing. Models and Applications: 17th International Conference, ICONIP 2010, Sydney, Australia, November 22-25, 2010, Proceedings, Part II 17, Springer, 2010, pp.~152--159.

\bibitem{jin2023optimal}
{\sc L.~Jin, D.~Lang, and N.~Lei}, {\em An optimal transport view of class-imbalanced visual recognition}, International Journal of Computer Vision, 131 (2023), pp.~2845--2863.

\bibitem{krawczyk2016learning}
{\sc B.~Krawczyk}, {\em Learning from imbalanced data: open challenges and future directions}, Progress in Artificial Intelligence, 5 (2016), pp.~221--232.

\bibitem{9848833}
{\sc M.~Li, Y.-M. Cheung, and Z.~Hu}, {\em Key point sensitive loss for long-tailed visual recognition}, IEEE Transactions on Pattern Analysis and Machine Intelligence, 45 (2023), pp.~4812--4825.

\bibitem{lin2017focal}
{\sc T.-Y. Lin, P.~Goyal, R.~Girshick, K.~He, and P.~Doll{\'a}r}, {\em Focal loss for dense object detection}, in Proceedings of the IEEE International Conference on Computer Vision, 2017, pp.~2980--2988.

\bibitem{liu2023data}
{\sc K.~Liu, H.~Zhang, Z.~Hu, F.~Wang, and P.~S. Yu}, {\em Data augmentation for supervised graph outlier detection with latent diffusion models}, arXiv preprint arXiv:2312.17679,  (2023).

\bibitem{liu2020deep}
{\sc Y.~Liu, A.~Jain, C.~Eng, D.~H. Way, K.~Lee, P.~Bui, K.~Kanada, G.~de~Oliveira~Marinho, J.~Gallegos, S.~Gabriele, et~al.}, {\em A deep learning system for differential diagnosis of skin diseases}, Nature Medicine, 26 (2020), pp.~900--908.

\bibitem{liu2021swin}
{\sc Z.~Liu, Y.~Lin, Y.~Cao, H.~Hu, Y.~Wei, Z.~Zhang, S.~Lin, and B.~Guo}, {\em Swin transformer: Hierarchical vision transformer using shifted windows}, in Proceedings of the IEEE/CVF International Conference on Computer Vision, 2021, pp.~10012--10022.

\bibitem{liu2019large}
{\sc Z.~Liu, Z.~Miao, X.~Zhan, J.~Wang, B.~Gong, and S.~X. Yu}, {\em Large-scale long-tailed recognition in an open world}, in Proceedings of the IEEE/CVF Conference on Computer Vision and Pattern Recognition, 2019, pp.~2537--2546.

\bibitem{ma2024geometric}
{\sc Y.~Ma, L.~Jiao, F.~Liu, S.~Yang, X.~Liu, and P.~Chen}, {\em Geometric prior guided feature representation learning for long-tailed classification}, International Journal of Computer Vision, 132 (2024), pp.~2493--2510.

\bibitem{marrakchi2021fighting}
{\sc Y.~Marrakchi, O.~Makansi, and T.~Brox}, {\em Fighting class imbalance with contrastive learning}, in International Conference on Medical Image Computing and Computer-Assisted Intervention, Springer, 2021, pp.~466--476.

\bibitem{nichol2021improved}
{\sc A.~Q. Nichol and P.~Dhariwal}, {\em Improved denoising diffusion probabilistic models}, in International Conference on Machine Learning, PMLR, 2021, pp.~8162--8171.

\bibitem{pacheco2020impact}
{\sc A.~G. Pacheco and R.~A. Krohling}, {\em The impact of patient clinical information on automated skin cancer detection}, Computers in Biology and Medicine, 116 (2020), p.~103545.

\bibitem{park2022majority}
{\sc S.~Park, Y.~Hong, B.~Heo, S.~Yun, and J.~Y. Choi}, {\em The majority can help the minority: Context-rich minority oversampling for long-tailed classification}, in Proceedings of the IEEE/CVF Conference on Computer Vision and Pattern Recognition, 2022, pp.~6887--6896.

\bibitem{qin2023class}
{\sc Y.~Qin, H.~Zheng, J.~Yao, M.~Zhou, and Y.~Zhang}, {\em Class-balancing diffusion models}, in Proceedings of the IEEE/CVF Conference on Computer Vision and Pattern Recognition, 2023, pp.~18434--18443.

\bibitem{ramesh2022hierarchical}
{\sc A.~Ramesh, P.~Dhariwal, A.~Nichol, C.~Chu, and M.~Chen}, {\em Hierarchical text-conditional image generation with clip latents}, arXiv preprint arXiv:2204.06125,  (2022).

\bibitem{rosenfeld2020a}
{\sc J.~S. Rosenfeld, A.~Rosenfeld, Y.~Belinkov, and N.~Shavit}, {\em A constructive prediction of the generalization error across scales}, in International Conference on Learning Representations, 2020, \url{https://openreview.net/forum?id=ryenvpEKDr}.

\bibitem{saharia2022photorealistic}
{\sc C.~Saharia, W.~Chan, S.~Saxena, L.~Li, J.~Whang, E.~L. Denton, K.~Ghasemipour, R.~Gontijo~Lopes, B.~Karagol~Ayan, T.~Salimans, et~al.}, {\em Photorealistic text-to-image diffusion models with deep language understanding}, Advances in Neural Information Processing Systems, 35 (2022), pp.~36479--36494.

\bibitem{salehi2017tversky}
{\sc S.~S.~M. Salehi, D.~Erdogmus, and A.~Gholipour}, {\em Tversky loss function for image segmentation using 3d fully convolutional deep networks}, in Machine Learning in Medical Imaging, Cham, 2017, Springer International Publishing, pp.~379--387.

\bibitem{Sandler_2018_CVPR}
{\sc M.~Sandler, A.~Howard, M.~Zhu, A.~Zhmoginov, and L.-C. Chen}, {\em Mobilenetv2: Inverted residuals and linear bottlenecks}, in Proceedings of the IEEE conference on computer vision and pattern recognition, 2018, pp.~4510--4520.

\bibitem{shen2015long}
{\sc T.~Shen, A.~Lee, C.~Shen, and C.~J. Lin}, {\em The long tail and rare disease research: the impact of next-generation sequencing for rare mendelian disorders}, Genetics Research, 97 (2015), p.~e15.

\bibitem{shrivastava2016training}
{\sc A.~Shrivastava, A.~Gupta, and R.~Girshick}, {\em Training region-based object detectors with online hard example mining}, in Proceedings of the IEEE Conference on Computer Vision and Pattern Recognition, 2016, pp.~761--769.

\bibitem{sinha2022class}
{\sc S.~Sinha, H.~Ohashi, and K.~Nakamura}, {\em Class-difficulty based methods for long-tailed visual recognition}, International Journal of Computer Vision, 130 (2022), pp.~2517--2531.

\bibitem{sohl2015deep}
{\sc J.~Sohl-Dickstein, E.~Weiss, N.~Maheswaranathan, and S.~Ganguli}, {\em Deep unsupervised learning using nonequilibrium thermodynamics}, in International Conference on Machine Learning, PMLR, 2015, pp.~2256--2265.

\bibitem{song2019generative}
{\sc Y.~Song and S.~Ermon}, {\em Generative modeling by estimating gradients of the data distribution}, Advances in Neural Information Processing Systems, 32 (2019).

\bibitem{szegedy2017inception}
{\sc C.~Szegedy, S.~Ioffe, V.~Vanhoucke, and A.~Alemi}, {\em Inception-v4, inception-resnet and the impact of residual connections on learning}, in Proceedings of the AAAI Conference on Artificial Intelligence, vol.~31, 2017, pp.~4278--4284.

\bibitem{pmlr-v97-tan19a}
{\sc M.~Tan and Q.~Le}, {\em Efficientnet: Rethinking model scaling for convolutional neural networks}, in Proceedings of the 36th International Conference on Machine Learning, vol.~97, 2019, pp.~6105--6114.

\bibitem{tschandl2018ham10000}
{\sc P.~Tschandl, C.~Rosendahl, and H.~Kittler}, {\em The ham10000 dataset, a large collection of multi-source dermatoscopic images of common pigmented skin lesions}, Scientific Data, 5 (2018), pp.~1--9.

\bibitem{wang2022pvt}
{\sc W.~Wang, E.~Xie, X.~Li, D.-P. Fan, K.~Song, D.~Liang, T.~Lu, P.~Luo, and L.~Shao}, {\em Pvt v2: Improved baselines with pyramid vision transformer}, Computational Visual Media, 8 (2022), pp.~415--424.

\bibitem{wang2017learning}
{\sc Y.-X. Wang, D.~Ramanan, and M.~Hebert}, {\em Learning to model the tail}, in Proceedings of the 31st International Conference on Neural Information Processing Systems, 2017, pp.~7032--7042.

\bibitem{ward2024crowdsourcing}
{\sc A.~Ward, J.~Li, J.~Wang, S.~Lakshminarasimhan, A.~Carrick, B.~Campana, J.~Hartford, T.~Tiyasirichokchai, S.~Virmani, R.~Wong, et~al.}, {\em Crowdsourcing dermatology images with google search ads: Creating a real-world skin condition dataset}, arXiv preprint arXiv:2402.18545,  (2024).

\bibitem{wu2024medsegdiff}
{\sc J.~Wu, R.~Fu, H.~Fang, Y.~Zhang, Y.~Yang, H.~Xiong, H.~Liu, and Y.~Xu}, {\em Medsegdiff: Medical image segmentation with diffusion probabilistic model}, in Medical Imaging with Deep Learning, PMLR, 2024, pp.~1623--1639.

\bibitem{wu2024medsegdiffv2}
{\sc J.~Wu, W.~Ji, H.~Fu, M.~Xu, Y.~Jin, and Y.~Xu}, {\em Medsegdiff-v2: Diffusion-based medical image segmentation with transformer}, in Proceedings of the AAAI Conference on Artificial Intelligence, vol.~38, 2024, pp.~6030--6038.

\bibitem{yang2023diffmic}
{\sc Y.~Yang, H.~Fu, A.~I. Aviles-Rivero, C.-B. Sch{\"o}nlieb, and L.~Zhu}, {\em Diffmic: Dual-guidance diffusion network for medical image classification}, in International Conference on Medical Image Computing and Computer-Assisted Intervention, Springer, 2023, pp.~95--105.

\bibitem{yang2020rethinking}
{\sc Y.~Yang and Z.~Xu}, {\em Rethinking the value of labels for improving class-imbalanced learning}, Advances in Neural Information Processing Systems, 33 (2020), pp.~19290--19301.

\bibitem{yang2022proco}
{\sc Z.~Yang, J.~Pan, Y.~Yang, X.~Shi, H.-Y. Zhou, Z.~Zhang, and C.~Bian}, {\em Proco: Prototype-aware contrastive learning for long-tailed medical image classification}, in International Conference on Medical Image Computing and Computer-Assisted Intervention, Springer, 2022, pp.~173--182.

\bibitem{9502525}
{\sc Z.~Yang, Q.~Xu, S.~Bao, X.~Cao, and Q.~Huang}, {\em Learning with multiclass auc: Theory and algorithms}, IEEE Transactions on Pattern Analysis and Machine Intelligence, 44 (2022), pp.~7747--7763.

\bibitem{zhang2017range}
{\sc X.~Zhang, Z.~Fang, Y.~Wen, Z.~Li, and Y.~Qiao}, {\em Range loss for deep face recognition with long-tailed training data}, in Proceedings of the IEEE International Conference on Computer Vision, 2017, pp.~5409--5418.

\bibitem{zhou2020bbn}
{\sc B.~Zhou, Q.~Cui, X.-S. Wei, and Z.-M. Chen}, {\em Bbn: Bilateral-branch network with cumulative learning for long-tailed visual recognition}, in Proceedings of the IEEE/CVF Conference on Computer Vision and Pattern Recognition, 2020, pp.~9719--9728.

\bibitem{zhou2023novel}
{\sc Y.-J. Zhou, W.~Liu, Y.~Gao, J.~Xu, L.~Lu, Y.~Duan, H.~Cheng, N.~Jin, X.~Man, S.~Zhao, et~al.}, {\em A novel multi-task model imitating dermatologists for accurate differential diagnosis of skin diseases in clinical images}, in International Conference on Medical Image Computing and Computer-Assisted Intervention, Springer, 2023, pp.~202--212.

\end{thebibliography}
\end{document}


\maketitle

\section{A detailed example}

Here we include some equations and theorem-like environments to show
how these are labeled in a supplement and can be referenced from the
main text.
Consider the following equation:
\begin{equation}
  
  a^2 + b^2 = c^2.
\end{equation}
You can also reference equations such as \cref{eq:matrices,eq:bb} 
from the main article in this supplement.

\lipsum[100-101]

\begin{theorem}
  An example theorem.
\end{theorem}

\lipsum[102]
 
\begin{lemma}
  An example lemma.
\end{lemma}

\lipsum[103-105]

Here is an example citation: \cite{KoMa14}.

\section[Proof of Thm]{Proof of \cref{thm:bigthm}}

\lipsum[106-112]

\section{Additional experimental results}
\Cref{tab:foo} shows additional
supporting evidence. 

\begin{table}[htbp]
{\footnotesize
  \caption{Example table} 
\begin{center}
  \begin{tabular}{|c|c|c|} \hline
   Species & \bf Mean & \bf Std.~Dev. \\ \hline
    1 & 3.4 & 1.2 \\
    2 & 5.4 & 0.6 \\ \hline
  \end{tabular}
\end{center}
}
\end{table}

\bibliographystyle{siamplain}
\bibliography{references}